\documentclass[11pt,a4paper]{article}
\usepackage[utf8]{inputenc}
\usepackage[T1]{fontenc}
\usepackage{amsmath,amssymb,amsfonts}
\usepackage{graphicx}
\usepackage{booktabs}
\usepackage{float}
\usepackage{geometry}
\usepackage{caption}
\usepackage{subcaption}
\usepackage{bm}
\usepackage[authoryear,round]{natbib}      
\usepackage[colorlinks=true, citecolor=blue, linkcolor=black, urlcolor=blue]{hyperref}
\title{Mixture of Experts for Low-Resource LLMs}
\author{
    \textbf{Ori Bar Joseph}\thanks{These authors contributed equally to this work.} \and
    \textbf{Smadar Arvatz}\footnotemark[1] \and
    {Noam Kayzer} \and
    {Dan Revital} \and
    \textbf{Sarel Weinberger}\thanks{Corresponding author}
}
\date{May 1, 2026}

\begin{document}
\maketitle

\begin{abstract}
Mixture-of-Experts (MoE) architectures have emerged as a dominant paradigm for scaling large language models (LLMs), yet expert routing behavior across languages, particularly underrepresented ones, remains poorly understood. We present a systematic analysis of expert routing dynamics in two architecturally distinct MoE models: a pure Transformer-based model (Qwen3-30B-A3B) and a hybrid Mamba-Transformer model (Nemotron-3-Nano-30B-A3B), using Hebrew as a morphologically rich, low-resource language. Our analysis reveals several key findings. First, pre-trained base models exhibit a pronounced deep-layer routing collapse for Hebrew, where usage entropy drops sharply in the final layers and tokens are increasingly concentrated among a narrow subset of experts, a pattern largely absent for English. Second, fine-tuning on data that better represents the target language substantially mitigates this imbalance: expert utilization entropy increases, the deep-layer collapse is reduced, and expert selection shifts from language-specific routing toward shared bilingual experts driven by functional specialization rather than language identity. To evaluate the generality of these findings across architectures and languages, we extend the analysis to Japanese using a fixed-architecture comparison. Deep-layer routing collapse is observed in Japanese with a profile quantitatively consistent with Hebrew, providing evidence that this pattern may generalize across typologically unrelated low-resource languages. Furthermore, within this controlled Japanese comparison, continual pre-training (CPT) on balanced data substantially corrects the collapse and promotes bilingual expert sharing, while supervised fine-tuning (SFT) alone leaves the deepest-layer imbalance largely unremediated. These routing improvements are associated with consistent downstream benchmark gains across both languages. These findings identify internal routing imbalance as a measurable and potentially correctable diagnostic marker for low-representation language performance in MoE systems, and suggest that routing entropy and expert specialization patterns offer a principled lens for understanding multilingual capacity allocation.

\end{abstract}

\section{Introduction}
\label{sec:introduction}
Mixture-of-Experts (MoE) architectures have become a central paradigm for scaling large language models, enabling substantial increases in model capacity with only minor losses in computational efficiency \citep{shazeer2017, fedus2021}. By routing each input token through a sparse subset of specialized feed-forward networks, MoE models increase total capacity while keeping per-token active computation relatively low, though often at the cost of greater memory footprint and serving complexity. This property has driven their adoption in several state-of-the-art systems, including Switch Transformers \citep{fedus2021}, GShard \citep{lepikhin2020}, GLaM \citep{du2022}, Mixtral \citep{jiang2024}, and DeepSeek-V2 \citep{deepseekv2_2024}.
The behavior of MoE routing mechanisms has primarily been studied through training stability and load balancing. Early work showed that without explicit regularization, routing can collapse into degenerate states in which a small subset of experts receives most tokens while others remain underutilized \citep{shazeer2017}. Subsequent work on scaling MoE models further identified training instability as a central challenge and proposed architectural solutions to address it \citep{fedus2021}. A number of solutions were proposed for this problem including the use of load-balancing losses in training and noisy top-k gating, as well as expert-choice routing, where experts select tokens instead of the reverse which leads to improved balance \citep{zhou2022}. Uneven expert utilization is not merely an efficiency concern but can also impact model quality \citep{zhou2022}. 

For large-scale multilingual MoE systems, prior work such as 
GShard \citep{lepikhin2020} demonstrates that routing behavior 
varies across inputs in a multilingual translation setting, though 
expert behavior has not been systematically characterized along 
linguistic dimensions. Studies in vision MoE models demonstrate that routing entropy provides a useful signal for characterizing expert utilization and specialization patterns \citep{riquelme2021}. More generally, token distribution across experts has been shown to be a 
critical factor for training stability and effective capacity utilization 
\citep{fedus2021, zoph2022}.

However, a key question remains underexplored: how do MoE routing systems behave for low-resource languages that receive substantially less pre-training exposure than dominant languages such as English? This is an important question for many lower-representation languages, as models for such languages are trained on highly imbalanced multilingual data distributions that strongly favor high-resource languages such as English \citep{joshi2020}. While multilingual pre-training enables cross-lingual transfer \citep{conneau2020, xue2021}, it remains unclear whether MoE routing mechanisms allocate computational capacity equitably across languages or whether pre-training imbalances induce systematic routing biases. Early evidence from GShard suggests that experts can exhibit emergent specialization patterns, though these have not been explicitly characterized along linguistic dimensions \citep{lepikhin2020}. Multi-task learning studies show that the balance between expert specialization and sharing affects performance on low-resource tasks \citep{ma2018}, a pattern we observe in the multilingual routing context. However, a direct, fine-grained analysis of per-language routing dynamics in modern large-scale MoE models is still missing.

Recent work has started to characterize these cross-lingual routing dynamics \citep{bandarkar2026}, showing that routing decisions have greater variance between languages in the first and final layers but exhibit more similarity in the middle layers. Critically, this similarity is less pronounced for lower-representation languages, which show more distinct routing patterns across all layers, including the middle ones. Our analysis corroborates this and further demonstrates that fine-tuning for the low-resource language significantly closes this gap, especially in the model's deep layers. 
To assess whether this collapse is specific to Hebrew or reflects a broader consequence of pre-training underrepresentation, we extend our analysis to Japanese, a typologically unrelated language differing from Hebrew in script, morphology, syntactic structure, and writing direction, using a controlled fixed-architecture comparison across three training variants. The similar collapse signatures observed across both languages provide cross-linguistic evidence that deep-layer routing imbalance may be a recurring property of MoE systems trained on imbalanced multilingual corpora, rather than an artifact of any particular language's structure.
We find that these routing changes are associated with improved downstream benchmark performance: CPT-induced routing reorganization aligns with more consistent NLP gains, while SFT, which corrects deep-layer collapse less completely than CPT in the comparisons studied, produces uneven results reflecting a redistribution rather than a broadening of model competencies.

In this work, we address this gap by focusing on Hebrew, a morphologically rich, low-resource language that is substantially underrepresented in typical pre-training corpora. We analyze two models of comparable scale but fundamentally different architectures: Qwen3-30B-A3B \citep{alibaba2025qwen3}, a pure Transformer-based MoE, and Nemotron-3-Nano-30B-A3B \citep{nvidia2025nemotron}, a hybrid Mamba-Transformer MoE. This comparison allows us to examine whether routing imbalance appears across distinct architectural families, while treating architecture-level conclusions as suggestive due to confounds in pre-training data, routing design, and expert configuration.
To study routing behavior, we introduce a set of complementary metrics: usage entropy, the Gini coefficient, the Language Specificity Index (LSI), layer-wise cosine similarity,selection entropy, Top-K expert overlap and rank correlation (Spearman and Kendall's $\tau$) (See Supplementary Materials), each capturing a distinct dimension of expert utilization and cross-lingual routing behavior.
Our analysis yields four main findings. First, both pre-trained models exhibit a pronounced deep-layer routing collapse for Hebrew: in the final layers, Hebrew tokens are routed to a narrow subset of experts, while English tokens are distributed more evenly. Second, continual pre-training (CPT) on a balanced Hebrew–English corpus substantially mitigates this imbalance, increasing entropy, reducing Gini, and shifting routing toward more language-agnostic behavior. Third, CPT leads to important differences between the two architectures: for the hybrid Mamba-Transformer model it largely eliminates deep-layer collapse while the pure Transformer model retains small residual imbalance. Fourth, we find that CPT is more consistently associated with remediation of deep-layer routing collapse than SFT in the comparisons studied. This suggests that the broader data exposure in CPT may facilitate a more substantial reconfiguration of the model's expert utilization than alignment-based tuning alone.
These disparities in routing behavior and downstream performance provide critical insights for researchers and engineers seeking to optimize Large Language Models (LLMs) for low-resource languages.

\section{Methods}
\label{sec:methods}

\subsection{Models}
\label{sec:models}

We selected two large language models based on the Mixture of Experts (MoE) architecture \citep{shazeer2017, fedus2021}, both sharing a comparable total parameter count (\textasciitilde30B) but with significantly fewer active parameters per forward pass (\textasciitilde3B). The low active parameter count also makes fine-tuning feasible under standard GPU constraints. The two models differ fundamentally in their architectural design, allowing us to investigate how routing imbalance appears across distinct MoE model families while keeping architecture-level causal claims appropriately cautious.

\subsubsection{Qwen3-30B-A3B-Base}
\label{sec:qwen}

Qwen3-30B-A3B-Base\citep{alibaba2025qwen3} is a pretrained causal language model. The model comprises 30.5 billion total parameters, of which only 3.3 billion are activated per token during inference. Its MoE architecture consists of 48 layers, each containing 128 routed experts with a top-8 routing strategy (i.e., 8 experts are activated per token). The attention mechanism employs Grouped Query Attention (GQA) \citep{ainslie2023} with 32 query heads and 4 key-value heads, and the maximum context length is 32,768 tokens. Notably, Qwen3 was pre-trained on approximately 36 trillion tokens spanning 119 languages (including Hebrew). The pre-training corpus comprises a diverse mix of high-quality data, including code, STEM, reasoning, books, multilingual, and synthetic data \citep{alibaba2025qwen3}.

\subsubsection{NVIDIA Nemotron-3-Nano-30B-A3B-Base-BF16}
\label{sec:nemotron}

NVIDIA Nemotron-3-Nano-30B-A3B-Base-BF16 \citep{nvidia2025nemotron} is a hybrid architecture that combines Mamba-2 state space layers with Transformer-based MoE and attention layers \citep{dao2024}. The model contains 31.6 billion total parameters, with 3.5 billion active parameters per token. The architecture is composed of 52 layers in total: 23 Mamba-2 layers, 23 MoE layers, and 6 attention layers employing Grouped Query Attention (GQA) \citep{ainslie2023}. Each MoE layer contains 128 routed experts plus 1 shared expert (which is activated for all tokens). The shared expert is activated for every token regardless of routing decisions, effectively providing a residual computation path common to all inputs. In total, 6 experts are activated per token in each MoE layer. Mamba-2 is a structured state space model designed for efficient long-sequence processing; the hybrid architecture pairs it with attention layers to retain precise contextual modeling. The model was pre-trained on approximately 25 trillion tokens. Notably, while the pre-training corpus included 15 primary non-English languages sourced from Common Crawl, Hebrew was included only as one of four additional languages with data drawn exclusively from Wikipedia and FineWeb-2 \citep{penedo2025}, suggesting substantially lower Hebrew representation.

\subsubsection{Model Comparison}
\label{sec:model_comparison}

Table~\ref{tab:model_comparison} summarizes the key architectural differences between the two models.

\begin{table}[htbp]
\centering
\caption{Architectural comparison of the two MoE models used in this study.}
\label{tab:model_comparison}
\begin{tabular}{lcc}
\toprule
\textbf{Property} & \textbf{Qwen3-30B-A3B} & \textbf{Nemotron-3-Nano-30B-A3B} \\
\midrule
Total Parameters & 30.5B & 31.6B\\
Active Parameters & 3.3B & 3.5B \\
Number of Layers & 48 & 52 (23 Mamba-2 + 23 MoE + 6 Attention) \\
Number of Experts & 128 & 128 + 1 shared \\
Active Experts per Token & 8 & 6 (top 5 routed + 1 shared) \\
Architecture & Transformer MoE & Mamba2-Transformer Hybrid MoE \\
Pre-training Tokens & \textasciitilde36T & 25T \\
Pre-training Languages & 119 (incl.\ Hebrew) & 20 (incl.\ Hebrew) \\
\bottomrule
\end{tabular}
\end{table}

The selection of these two models allows for a comparative analysis across two architecturally distinct model families: both models share similar scale in terms of total and active parameters, though they differ in their computational paradigm. Qwen3-30B-A3B-Base is a pure Transformer-based MoE \citep{shazeer2017, fedus2021}  whereas Nemotron-3-Nano-30B-A3B-Base-BF16 \citep{nvidia2025nemotron} is a hybrid Mamba-Transformer MoE \citep{dao2024}. This contrast enables us to examine whether the underlying architecture may influence the effectiveness of fine-tuning for a morphologically rich, low-resource language such as Hebrew \citep{tsarfaty2020}. Because the models also differ in pre-training corpus, routing configuration, number of MoE layers, active experts per token, and the presence of a shared expert, we treat architecture-level interpretations as hypotheses rather than causal conclusions.

\subsection{Training Data}
\label{sec:training_data}

The training corpus was constructed with a deliberate language ratio of approximately 60\% Hebrew and 40\% English. This composition was designed to expose the models to a substantial volume of Hebrew data, which is significantly underrepresented  in the pre-training corpora of most large language models \citep{lauscher2020, tsarfaty2020}. At the same time, this ratio was chosen to preserve the models' English capabilities and mitigate catastrophic forgetting of knowledge acquired during pre-training \citep{kirkpatrick2017, gururangan2020}. The inclusion of English data serves as an anchor to maintain cross-lingual transfer capabilities \citep{conneau2019} and prevents degradation of general-purpose performance.

\subsection{Training Strategies}
\label{sec:training_strategies}
The CPT variants of both models were trained using a shared two-stage pipeline: the base model 
first underwent continual pre-training (CPT) \citep{gururangan2020} on the 
Hebrew-enriched corpus with a global batch size of 2048 for short-context training, followed by a global batch size of 256 for long-context training (sequence length 65K tokens). In addition, a second Nemotron variant (instruct64) 
was trained using supervised fine-tuning (SFT) directly on the Hebrew-enriched 
corpus, providing a complementary perspective on the relative contribution of 
each training stage to routing reorganization and downstream performance.

\subsection{Expert Load Balancing}
\label{sec:load_balancing}

A critical hyperparameter in the training of Mixture-of-Experts (MoE) models 
is the auxiliary load balancing loss coefficient ($\alpha$), which governs the 
distribution of computational load across experts. Setting $\alpha$ too high 
risks overriding the task-specific loss signal and causing catastrophic forgetting 
of knowledge acquired during pre-training. Conversely, setting $\alpha$ too low 
may lead to expert collapse, a degenerate state where a small subset of experts 
bears the majority of the computational load while the remaining experts are 
underutilized \citep{shazeer2017, fedus2021}. To select an appropriate value for 
$\alpha$, we compared three candidate settings: $\alpha = 0$, the commonly used 
default of $\alpha = 0.02$, and a reduced value of $\alpha = 0.002$.

\begin{description}
  \item[$\alpha = 0$:] Disabling the load-balancing signal led to substantially weaker routing adaptation in our ablation. Without routing pressure, expert assignments remained closer to the pre-trained configuration, reducing the model's ability to adapt effectively to the Hebrew-enriched corpus.
  
  \item[$\alpha = 0.02$:] This setting produced a pronounced loss spike early in 
  training followed by a slow and unstable recovery, as documented in the training 
  curves provided in the Supplementary Materials.
  
  \item[$\alpha = 0.002$:] This value yielded a smooth, monotonically decreasing 
  loss that converged to a lower final value, demonstrating more stable and 
  effective optimization (see Supplementary Materials for full training plots and 
  details).
\end{description}

Based on this three-way ablation, we selected $\alpha = 0.002$ for all 
experiments. This choice allows the routing mechanism to adapt to the new 
training distribution while remaining small enough to avoid destabilizing the 
task-specific loss signal. This balance is particularly vital for low-resource language
fine-tuning, where the model must acquire new linguistic patterns without 
erasing broad multilingual and general-purpose capabilities established during 
pre-training \citep{kirkpatrick2017, gururangan2020}.

\subsection{Evaluation}
\label{sec:evaluation}

Evaluation was conducted on a sample of 1300 examples drawn from the \textbf{ted\_he\_en\_chunks} dataset, which contains parallel Hebrew--English sentence pairs extracted from TED talk transcriptions \citep{cettolo2012}, and the \textbf{FLORES-200} benchmark \citep{nllb2022}, a high-quality evaluation dataset designed for multilingual machine translation.
This benchmark was selected for several reasons: (1) TED talks cover a broad range of topics, providing thematic diversity; (2) the parallel nature of the data enables direct assessment of cross-lingual transfer; and (3) the text represents natural, fluent discourse rather than synthetic or template-generated content, offering a realistic testbed for evaluating Hebrew language generation and comprehension capabilities.
We also conducted evaluation for Japanese using parallel English--Japanese sentence pairs from the TED dataset (same as Hebrew). This approach ensures that the assessment of cross-lingual transfer and expert specialization is consistent across both Hebrew and Japanese. The same filtering protocol was applied to exclude special tokens and focus solely on content tokens. Because TED talks represent a single discourse domain, we treat these routing measurements as a controlled diagnostic setting rather than a comprehensive estimate of routing behavior across all domains.

\subsection{Expert Routing Analysis}
\label{sec:routing_analysis}

\subsubsection{Analysis Protocol}
\label{sec:analysis_protocol}

To investigate how MoE models allocate computation across experts when processing Hebrew versus English, we conducted a systematic analysis of expert routing patterns at every MoE layer. Hebrew and English texts were processed as separate forward passes through the model; each text chunk from the \texttt{ted\_he\_en\_chunks} and the \textbf{FLORES-200} benchmark \citep{nllb2022} datasets were fed independently, ensuring that the routing decisions for each language were uncontaminated by the presence of the other language in the input context. Router logits were captured at each MoE layer using PyTorch forward hooks registered on the gate (router) modules. For each layer, the hook intercepts the hidden states entering the gate and computes the raw router logits via a linear projection (i.e., $\mathbf{z} = \mathbf{h} W_g^T$, where $\mathbf{h}$ is the hidden state and $W_g$ is the gate weight matrix). These logits are then converted to probability distributions via softmax normalization. From the resulting probability vectors, we derive both (1) the discrete top-K expert assignments and (2) the continuous routing probability scores for all experts. To ensure that the analysis reflects genuine linguistic processing rather than artifacts of input formatting, special tokens (padding tokens, beginning/end-of-sequence markers, chat template tokens) were excluded using a content mask. Only content tokens contributed to the computation of all reported metrics. For each model variant, the analysis was performed over 1300 parallel Hebrew--English text chunks (approximately 250-350 words per chunk per language), yielding approximately 100,000-160,000 content tokens per language. Since routing decisions are made independently at each MoE layer for each token, this produces several million routing observations per language, providing sufficient statistical power for robust characterization of expert utilization patterns.

All metrics were computed per MoE layer. For models containing non-MoE layers (e.g., the Mamba-2 and GQA attention layers in Nemotron), the routing analysis is restricted to the 23 MoE layers. Two types of aggregated outputs were retained for downstream analysis: (1) the mean softmax routing probability per expert per layer per language, and (2) the mean binary activation rate per expert per layer per language (i.e., the fraction of tokens for which each expert was among the top-K selected).

\paragraph{Controls and Robustness Checks.}
To distinguish language-driven routing effects from sampling noise and tokenization artifacts, we include several robustness checks. First, we verify that reported entropy, Gini, and cosine-similarity differences between languages exceed one standard deviation across text chunks in the deep layers where collapse is most pronounced, ensuring that observed gaps reflect stable signal rather than sampling noise. Second, we compare Hebrew–English and Japanese–English gaps against same-language split controls, in which each language is randomly divided into two subsets and analyzed as if they were distinct languages. Third, we report tokenization statistics across languages, including tokens per character and tokens per whitespace-delimited segment, to assess whether segmentation differences contribute to routing imbalance. Finally, we evaluate the sensitivity of LSI-based expert categorization to multiple thresholds rather than relying solely on the $\pm0.1$ cutoff.

\subsubsection{Expert Load Distribution Metrics}
\label{sec:load_metrics}

\paragraph{Usage Entropy.}
Usage entropy quantifies how uniformly the computational load is distributed across all experts within a given layer after the routing decisions have been made \citep{shannon1948}. For a layer with $N$ experts, let $p_i$ denote the fraction of tokens routed to expert $i$. The usage entropy is defined as:
\begin{equation}
H_{\text{usage}} = -\sum_{i=1}^{N} p_i \log_2 p_i
\end{equation}
where $$p_i = \frac{c_i}{\sum_{j=1}^{N} c_j}$$ and $c_i$ is the number of tokens assigned to expert $i$. The maximum entropy $$H_{\max} = \log_2 N$$ is achieved when all experts receive an equal share of the tokens ($p_i = \frac{1}{N}$ for all $i$). Values close to $H_{\max}$ indicate a well-balanced load distribution, while low values indicate that a small subset of experts dominates.

\paragraph{Gini Coefficient.}
The Gini coefficient, borrowed from economics where it measures income inequality \citep{gini1912}, is used here to quantify the inequality of workload distribution among experts. For a layer with $N$ experts whose token counts are sorted in ascending order as $c_1 \leq c_2 \leq \ldots \leq c_N$, the Gini coefficient is computed as:
\begin{equation}
G = \frac{\sum_{i=1}^{N}(2i - N - 1) \cdot c_i}{N \sum_{i=1}^{N} c_i}
\end{equation}
A Gini coefficient of 0 indicates perfect equality (all experts receive the same number of tokens), while a value approaching 1 indicates maximal inequality (a single expert processes all tokens). This metric complements usage entropy by providing an intuitive measure of load imbalance that is particularly sensitive to skewed distributions.

\subsubsection{Language Specialization Metrics}
\label{sec:specialization_metrics}

\paragraph{Language Specificity Index (LSI).}
To quantify the degree to which individual experts specialize in a particular language, we introduce a novel metric, the Language Specificity Index (LSI). For each expert $i$, let $a_i^{\text{he}}$ denote its mean activation score (router probability) when processing Hebrew tokens, and $a_i^{\text{en}}$ its mean activation score when processing English tokens. The LSI is defined as:
\begin{equation}
\text{LSI}_i = \frac{a_i^{\text{he}} - a_i^{\text{en}}}{a_i^{\text{he}} + a_i^{\text{en}}}
\end{equation}
The LSI ranges from $-1$ to $+1$, where:
\begin{itemize}
    \item $\text{LSI}_i > 0.1$: expert $i$ is classified as a \textbf{Hebrew-specialized} expert
    \item $\text{LSI}_i < -0.1$: expert $i$ is classified as an \textbf{English-specialized} expert
    \item $|\text{LSI}_i| \leq 0.1$: expert $i$ is classified as a \textbf{shared (language-agnostic)} expert
\end{itemize}
The threshold of $\pm 0.1$ was chosen to allow a neutral zone that accounts for minor stochastic variation in routing. To ensure that conclusions are not an artifact of this threshold, we perform a sensitivity analysis over alternative thresholds ($\pm0.05$, $\pm0.10$, $\pm0.15$, and $\pm0.20$) and report whether the shared-versus-language-specific expert trends remain stable. By computing the LSI for all experts within each layer, we obtain a complete picture of the language-specific allocation of model capacity.

\subsubsection{Cross-Lingual Similarity Metrics}
\label{sec:similarity_metrics}

\paragraph{Layer-wise Cosine Similarity.}
To measure how similarly each layer processes Hebrew and English, we compute the cosine similarity between the language-specific expert utilization vectors. For a given layer, let $\mathbf{u}^{\text{he}} \in \mathbb{R}^N$ be the vector of expert usage counts (or mean routing probabilities) for Hebrew tokens, and $\mathbf{u}^{\text{en}} \in \mathbb{R}^N$ the corresponding vector for English tokens. The cosine similarity is:
\begin{equation}
\text{cos\_sim} = \frac{\mathbf{u}^{\text{he}} \cdot \mathbf{u}^{\text{en}}}{\|\mathbf{u}^{\text{he}}\| \times \|\mathbf{u}^{\text{en}}\|}
\end{equation}
Values close to 1.0 indicate that the layer processes both languages using similar expert utilization patterns (language-agnostic layers), while values close to 0 indicate divergent processing (language-specific layers).

\noindent\textit{Note: All metrics are computed per MoE layer and reported both as layer-wise profiles and as summary statistics aggregated across layers. For models with non-MoE layers (e.g., the Mamba-2 and attention layers in Nemotron), the routing analysis is restricted to the MoE layers only.}

\section{Results}
\label{sec:results}

\subsection{Expert Usage Entropy}
\label{sec:usage_entropy}

Figure~\ref{fig:usage_entropy} presents the usage entropy across all MoE layers for each model variant, comparing Hebrew (red) and English (blue) token routing. The theoretical maximum entropy for 128 experts is $H_{\max} = \log_2 128 = 7.0$ bits, indicated by the dashed grey line. We organize our analysis along three axes: (1) architectural differences between the two base models, (2) the effect of fine-tuning within each architecture, and (3) a comparison of training strategies for the Nemotron model.

\subsubsection{Architectural Comparison: Base Models}
\label{sec:usage_arch}

A comparison of the two original (pre-trained) base models: Qwen3-30B-A3B-Base and Nemotron-3-Nano-30B-A3B-Base reveals a striking difference in the stability of expert utilization across layers.

In the \textbf{Qwen base model} (Figure~\ref{fig:usage_entropy}a), both languages exhibit relatively stable usage entropy throughout most of the 48 MoE layers, fluctuating in the range of approximately 5.0--5.5 bits.
This represents roughly 71–79\% of the theoretical maximum of 7.0 bits, indicating that even in well-balanced layers a meaningful subset of experts receives negligible token load. English consistently maintains higher entropy than Hebrew across the majority of layers, indicating a more uniform distribution of tokens across experts.

In contrast, the \textbf{Nemotron base model} (Figure~\ref{fig:usage_entropy}c) displays considerably greater variability across its 23 MoE layers, with overall entropy values that are notably higher than those observed in the Qwen base model, suggesting a more uniform token distribution across experts. The model displays heightened routing volatility, with layer-to-layer fluctuations that are substantially larger than those observed in the Qwen model. One possible explanation is that this instability reflects two complementary factors inherent to the Nemotron architecture. First, the Nemotron model is a hybrid architecture in which MoE layers are interleaved with Mamba-2 state space layers \citep{dao2024} and standard attention layers; since only the MoE layers participate in expert routing, the non-MoE layers that intervene between consecutive MoE layers may introduce discontinuities in the evolution of hidden state representations, leading to more erratic routing behavior. Second, the Mamba-2 layers, which process information through a structured state space mechanism rather than attention, may alter the distributional properties of the hidden states in ways that perturb the downstream router's ability to distribute tokens uniformly.

\subsubsection{Deep-Layer Entropy Collapse in Base Models}
\label{sec:usage_collapse}

A particularly notable finding, consistently observed across both base models, is a sharp decline in usage entropy in the deepest layers of the network. In the Qwen base model, this manifests as a pronounced drop in Hebrew entropy in the final layers, where the entropy gap widens dramatically with Hebrew entropy falling consistently well below that of English (Figure~\ref{fig:usage_entropy}a, panel b). The Nemotron base model exhibits an analogous pattern: Hebrew entropy collapses in the terminal MoE layers, with the gap bar chart (Figure~\ref{fig:usage_entropy}c, panel b) showing that the largest discrepancies between English and Hebrew are concentrated in the deepest layers, where English entropy substantially exceeds Hebrew entropy.

This deep-layer entropy collapse for Hebrew is highly characteristic of the behavior observed when MoE models encounter low-resource languages \citep{bandarkar2026} . In the deepest layers, where the network performs the most abstract and language-specific computations \citep{tenney2019}, a disproportionate fraction of Hebrew tokens are funneled through a narrow subset of experts. This concentration reflects the limited diversity of Hebrew representations acquired during pre-training: having seen far fewer Hebrew tokens than English tokens \citep{lauscher2020}, the model has developed specialized routing pathways for Hebrew that rely on a restricted expert pool rather than distributing the computational load broadly. The result is a form of partial expert collapse \citep{fedus2021} that is language-specific and depth-dependent: a degradation pattern consistent with the hypothesis that low representation during pre-training contributes to depth-dependent routing imbalance in MoE architectures.

\subsubsection{Effect of CPT: Qwen}
\label{sec:usage_qwen_ft}

A comparison between the original Qwen base model (Figure~\ref{fig:usage_entropy}a) and the CPT Qwen model (Figure~\ref{fig:usage_entropy}b) reveals a substantial improvement in Hebrew expert utilization following CPT on the Hebrew-enriched corpus.

In the CPT model, the Hebrew entropy curve (red) rises markedly across nearly all layers, and in fact \textbf{surpasses} the English entropy curve (blue) throughout much of the network. The entropy gap chart (Figure~\ref{fig:usage_entropy}b, panel b) confirms this reversal: the difference (English - Hebrew) is negative across the vast majority of layers, indicating that Hebrew tokens now achieve a more uniform distribution across experts than English tokens. This is a noteworthy inversion of the base model pattern, where English tokens had consistently higher routing entropy than Hebrew.

This improvement is consistent with the diversification of Hebrew routing pathways: after CPT, the router distributes Hebrew tokens across a broader set of experts rather than concentrating them in a narrow subset. This suggests that the model may have developed richer and more varied Hebrew representations that activate a wider expert repertoire.

However, the CPT does not entirely resolve the deep-layer entropy collapse observed in the base model. While the magnitude of the collapse is reduced, the final layers still exhibit a noticeable drop in Hebrew entropy relative to the preceding layers (Figure~\ref{fig:usage_entropy}b, panel b). This persistent vulnerability in the deepest layers suggests that the most abstract, language-specific computations remain partially constrained by the routing patterns established during pre-training: a residual effect that CPT on the pure Transformer (Qwen3) architecture alone may be 
insufficient to fully overcome, though as Section~\ref{sec:usage_nemotron_ft} 
demonstrates, the hybrid Nemotron model does not share this limitation.

\subsubsection{Effect of CPT: Nemotron}
\label{sec:usage_nemotron_ft}

\begin{figure}[H]
\centering
\begin{subfigure}[b]{0.48\textwidth}
    \includegraphics[width=\textwidth]{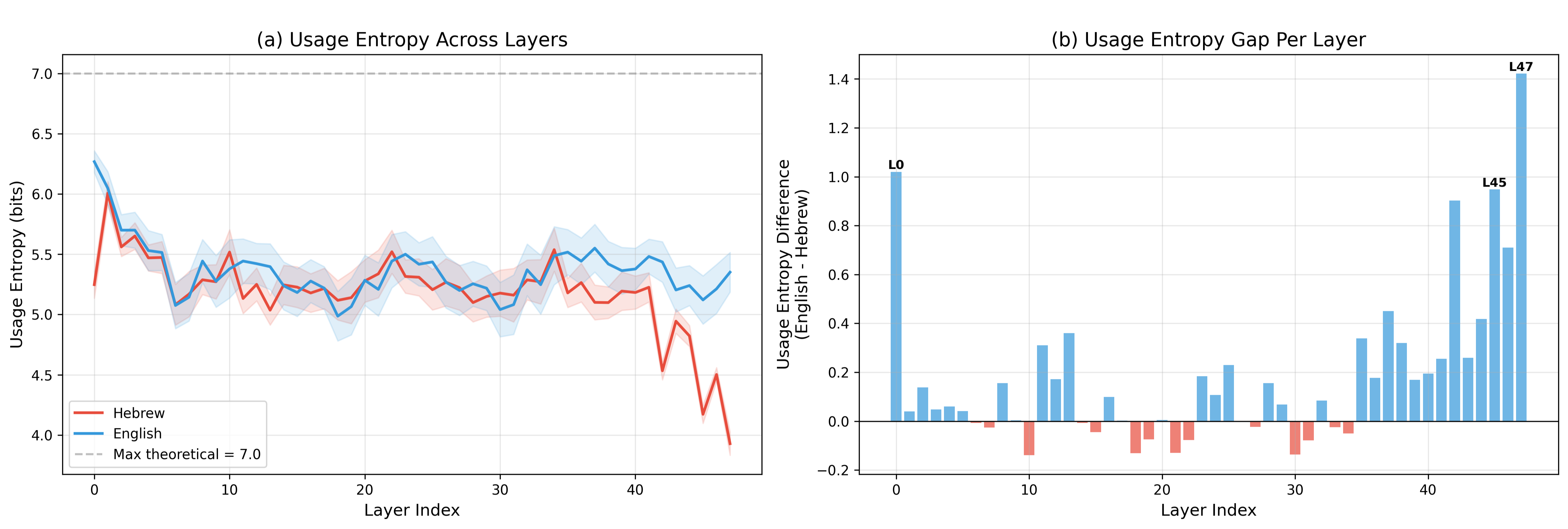}
    \caption{Qwen3 base}
\end{subfigure}
\hfill
\begin{subfigure}[b]{0.48\textwidth}
    \includegraphics[width=\textwidth]{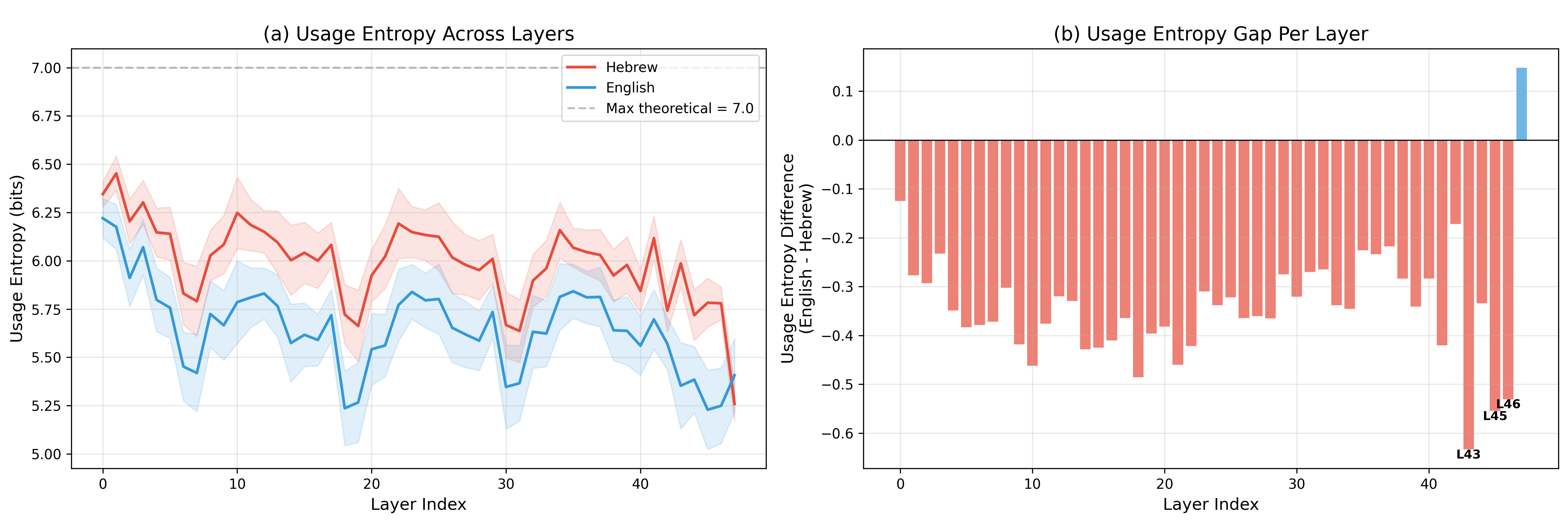}
    \caption{Qwen3 CPT}
\end{subfigure}
\\[0.5em]
\begin{subfigure}[b]{0.48\textwidth}
    \includegraphics[width=\textwidth]{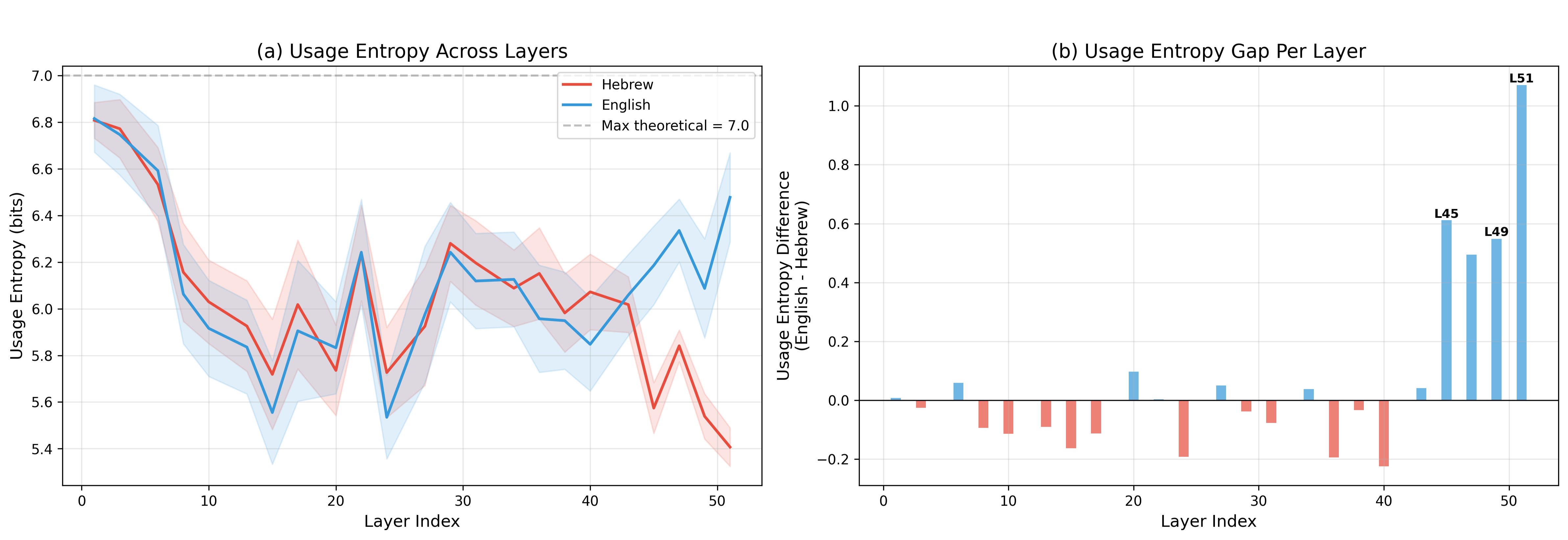}
    \caption{Nemotron base}
\end{subfigure}
\hfill
\begin{subfigure}[b]{0.48\textwidth}
    \includegraphics[width=\textwidth]{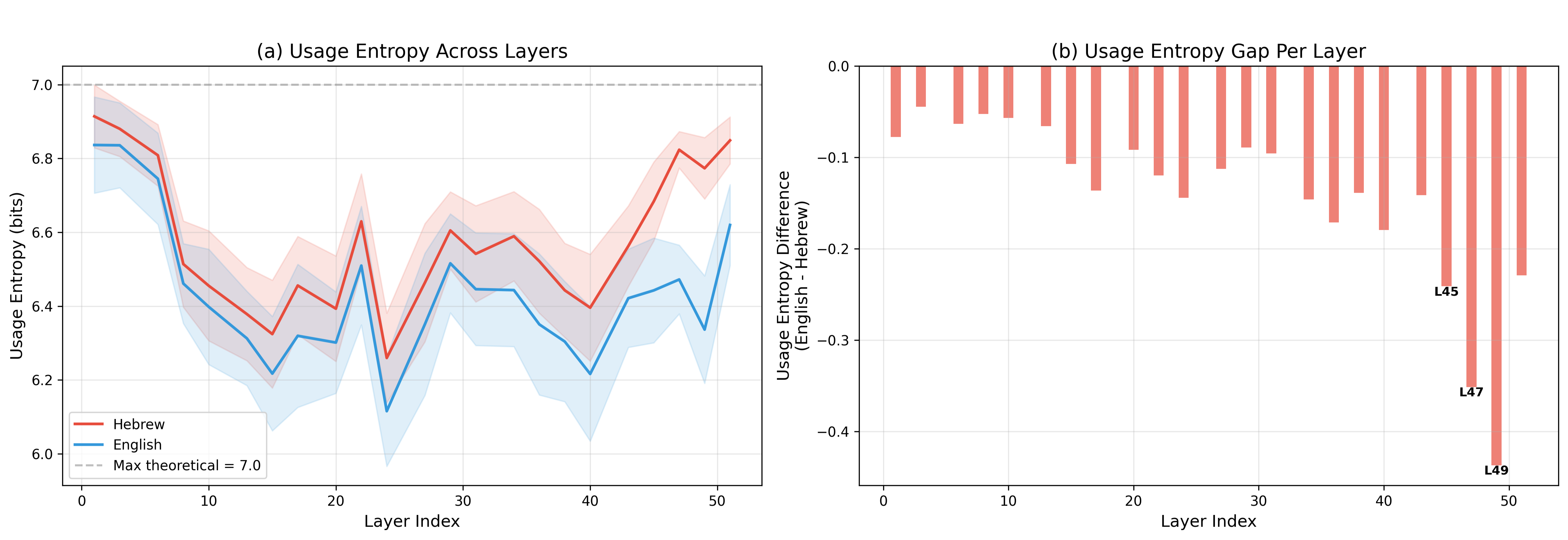}
    \caption{Nemotron base256}
\end{subfigure}
\\[0.5em]
\begin{subfigure}[b]{0.48\textwidth}
    \includegraphics[width=\textwidth]{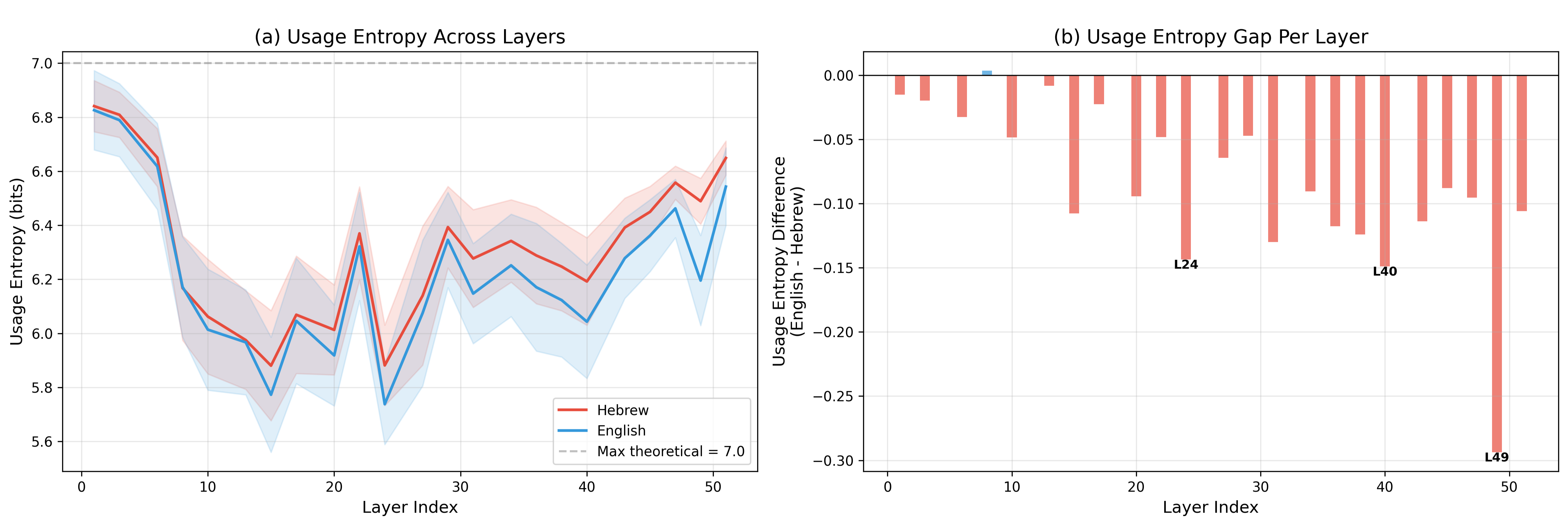}
    \caption{Nemotron instruct64}
\end{subfigure}
\caption{Usage entropy across MoE layers for all model variants. Each panel displays (a) the layer-wise usage entropy for Hebrew (red) and English (blue) with shaded $\pm 1$ standard deviation bands, and (b) the per-layer entropy gap (English $-$ Hebrew), where negative values indicate higher Hebrew entropy. The dashed grey line at 7.0 bits marks the theoretical maximum for 128 experts. Fine-tuning on Hebrew-enriched data consistently improves Hebrew usage entropy, but the magnitude and depth-profile of improvement depend on both the architecture and training strategy.}
\label{fig:usage_entropy}
\end{figure}

We compare two Nemotron variants: a CPT-trained model (base256; Figure~\ref{fig:usage_entropy}d) and a direct SFT model (instruct64; Figure~\ref{fig:usage_entropy}e). Both improve Hebrew expert utilization relative to the Nemotron base model (Figure~\ref{fig:usage_entropy}c), although CPT produces the larger absolute entropy increase.

Under both training configurations, Hebrew entropy (red) is elevated across all MoE layers relative to the base model, and consistently exceeds English entropy (blue) throughout the network. The entropy gap charts (Figures~\ref{fig:usage_entropy}d and \ref{fig:usage_entropy}e, panel b) confirm that the difference (English - Hebrew) is negative across nearly all layers in both variants, indicating that fine-tuning successfully rebalances the expert load distribution in favor of Hebrew regardless of the specific training strategy employed.

Critically, and in contrast to the CPT Qwen model, \textbf{both} Nemotron training strategies successfully mitigate the severe deep-layer entropy collapse observed in the base model. Rather than the sharp degradation seen in the final layers of the Nemotron base and the Qwen post CPT, both the base256 and instruct64 variants maintain relatively stable Hebrew entropy through the deepest layers, with the entropy curves exhibiting a recovery or plateau rather than a collapse. This suggests that the Nemotron training setup and architecture may be more amenable to correcting deep-layer imbalance, although this comparison remains confounded by differences in routing design, shared-expert structure, and pre-training data.

The primary distinction between the two Nemotron training strategies lies in the \textbf{magnitude} of entropy improvement. The base256 model, which underwent CPT \citep{gururangan2020}, achieves consistently higher absolute entropy values across all layers compared to the instruct64 model. This elevated baseline suggests that CPT facilitates a broader and more uniform utilization of the expert pool. We posit that the continual pre-training stage allows the model to develop richer internal representations of Hebrew in an unsupervised manner prior to instruction alignment, ultimately resulting in more diverse routing patterns. While the instruct64 model (the SFT-only Nemotron variant, introduced in Section~\ref{sec:training_strategies}) achieves qualitatively similar relative improvements, 
effectively correcting deep-layer collapse, it operates at a lower entropy baseline 
and does not match the broader expert utilization achieved by CPT.


\subsubsection{Summary}
\label{sec:usage_summary}

Across all model variants, three principal findings emerge from the usage entropy analysis. First, the two architectures differ fundamentally in their baseline routing stability: the pure Transformer-based Qwen model exhibits smooth, gradual entropy profiles, while the hybrid Mamba-Transformer Nemotron model displays greater layer-to-layer variability, likely a consequence of interleaving MoE layers with non-MoE computational blocks. Second, both base models exhibit a characteristic deep-layer entropy collapse for Hebrew, reflecting the limited diversity of Hebrew representations acquired during pre-training. Third, fine-tuning on a Hebrew-enriched corpus consistently elevates Hebrew usage entropy and reverses the language gap in favor of Hebrew across both architectures, but the tested model families differ in their response to fine-tuning in the deepest layers: the Nemotron model (under both training strategies) successfully corrects the deep-layer collapse, while the Qwen model retains a residual entropy degradation in its final layers.

\subsection{Gini Coefficient and Active Expert Count}
\label{sec:gini}

The Gini coefficient \citep{gini1912} provides a complementary perspective to usage entropy by quantifying the inequality of workload distribution among experts. While entropy is most sensitive to the effective number of utilized experts (particularly penalizing distributions with near-zero usage experts), the Gini coefficient directly measures pairwise inequality between experts and is thus more sensitive to the relative gaps in workload across the full distribution. The two metrics are generally inversely related, but they can diverge when, for example, a few experts are heavily overloaded while the remaining experts share the residual load relatively evenly (producing moderate entropy but a high Gini coefficient).

\textbf{Base models.} In both the Qwen and Nemotron base models, Hebrew and English exhibit broadly similar Gini profiles across most layers, with a notable divergence in the deepest layers where the Hebrew Gini coefficient rises sharply, mirroring the entropy collapse documented in Section~\ref{sec:usage_entropy}. The Nemotron base model operates at lower overall Gini values than the Qwen base, indicating a more uniform workload distribution across experts. However, the Nemotron active expert counts (panel c) display substantially greater layer-to-layer variability, consistent with the noisy entropy profiles observed earlier and attributable to the same architectural factors, namely, the interleaving of MoE layers with Mamba-2 \citep{dao2024}  and attention layers in the hybrid design. Despite this variability, the Nemotron base model tends to maintain a higher number of active experts across most layers than the Qwen base, suggesting that its routing mechanism engages a broader pool of experts even if the distribution among them is uneven.

\textbf{Effect of fine-tuning.} Fine-tuning on the Hebrew-enriched corpus reduces the Gini coefficient for Hebrew across both architectures, confirming that the improved entropy documented earlier corresponds to a genuinely more equitable distribution of tokens across experts, not merely a redistribution among already-active experts. In the CPT Qwen model, the Hebrew Gini decreases and the number of active experts increases relative to the base model, indicating that fine-tuning recruits previously underutilized experts into the Hebrew routing pathways. The same pattern is observed in both Nemotron fine-tuned variants: Hebrew Gini values decline and active expert counts rise. As with the entropy analysis, both Nemotron training strategies (base256 and instruct64) successfully prevent the sharp Gini increase in the deepest layers that characterizes the base model, whereas the CPT Qwen retains a residual deep-layer Gini elevation for Hebrew.
\begin{figure}[H]
\centering
\begin{subfigure}[b]{0.48\textwidth}
    \includegraphics[width=\textwidth]{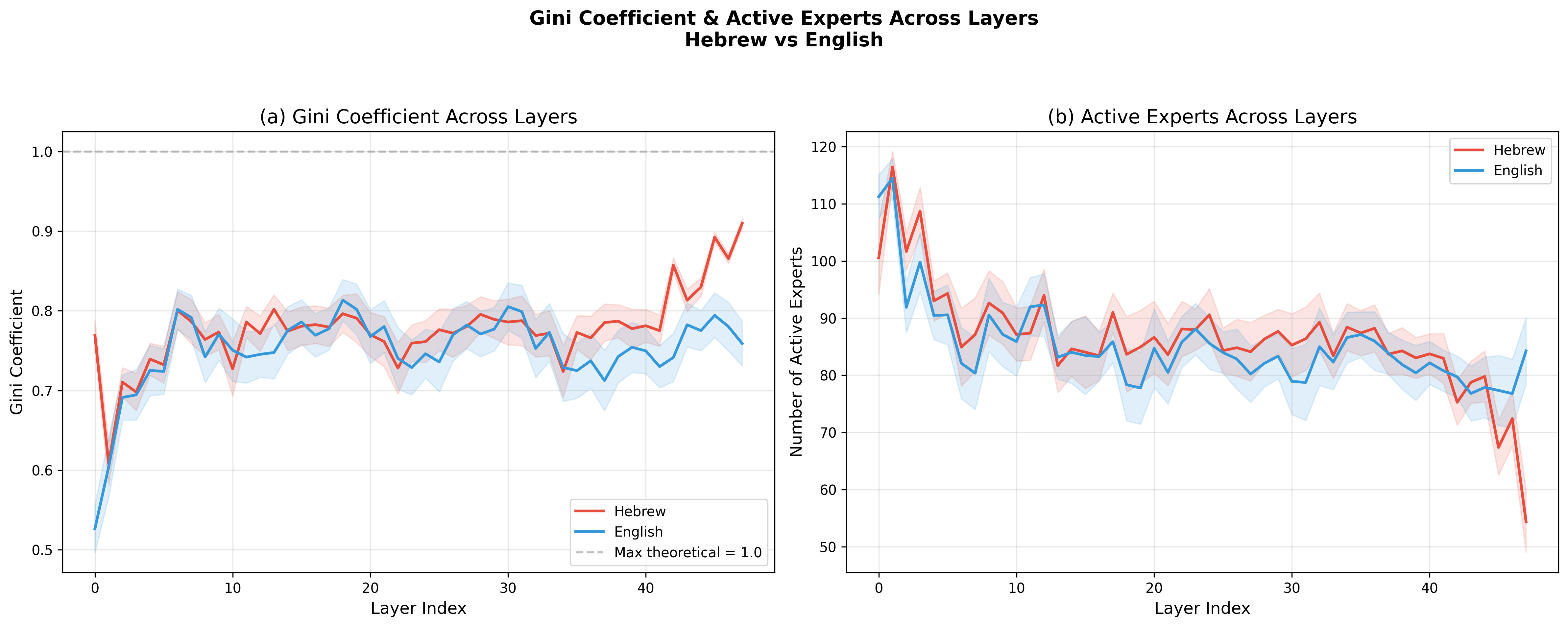}
    \caption{Qwen3 base}
\end{subfigure}
\hspace{2em}
\begin{subfigure}[b]{0.48\textwidth}
    \includegraphics[width=\textwidth]{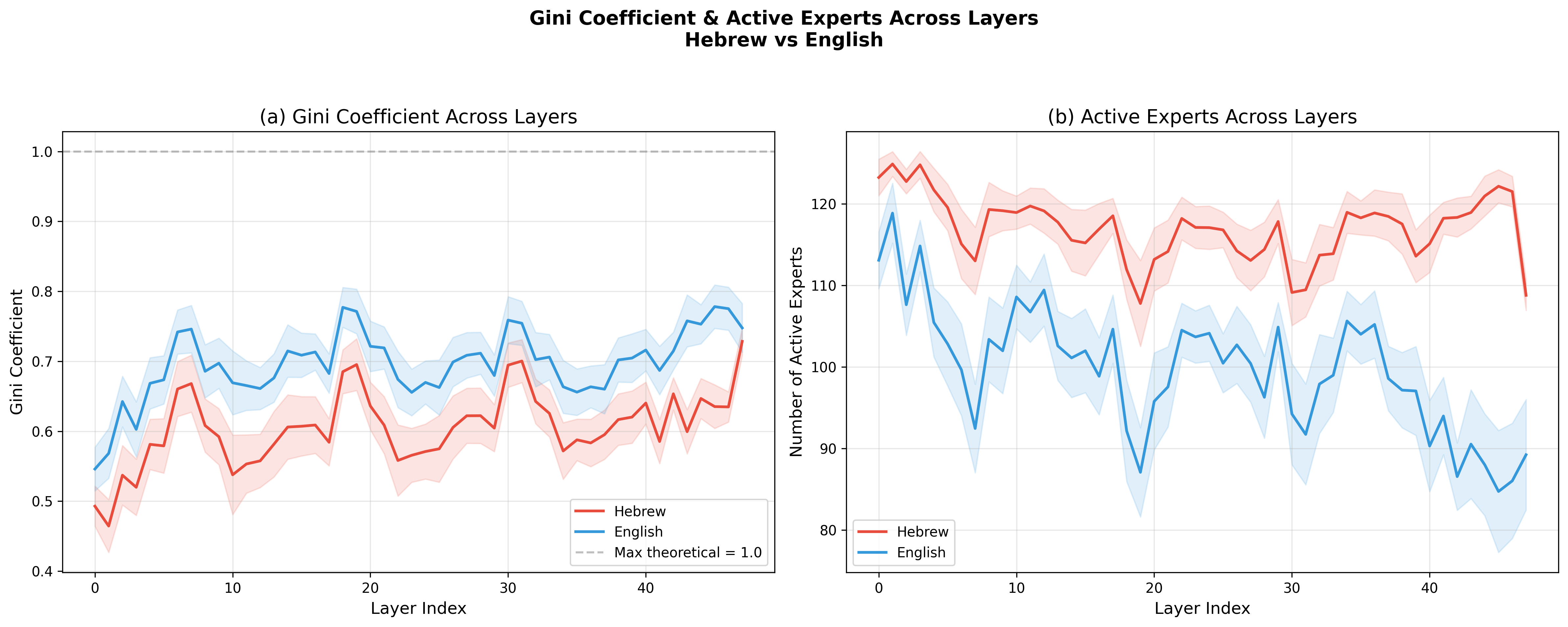}
    \caption{Qwen3 CPT}
\end{subfigure}
\\[0.5em]
\begin{subfigure}[b]{0.48\textwidth}
    \includegraphics[width=\textwidth]{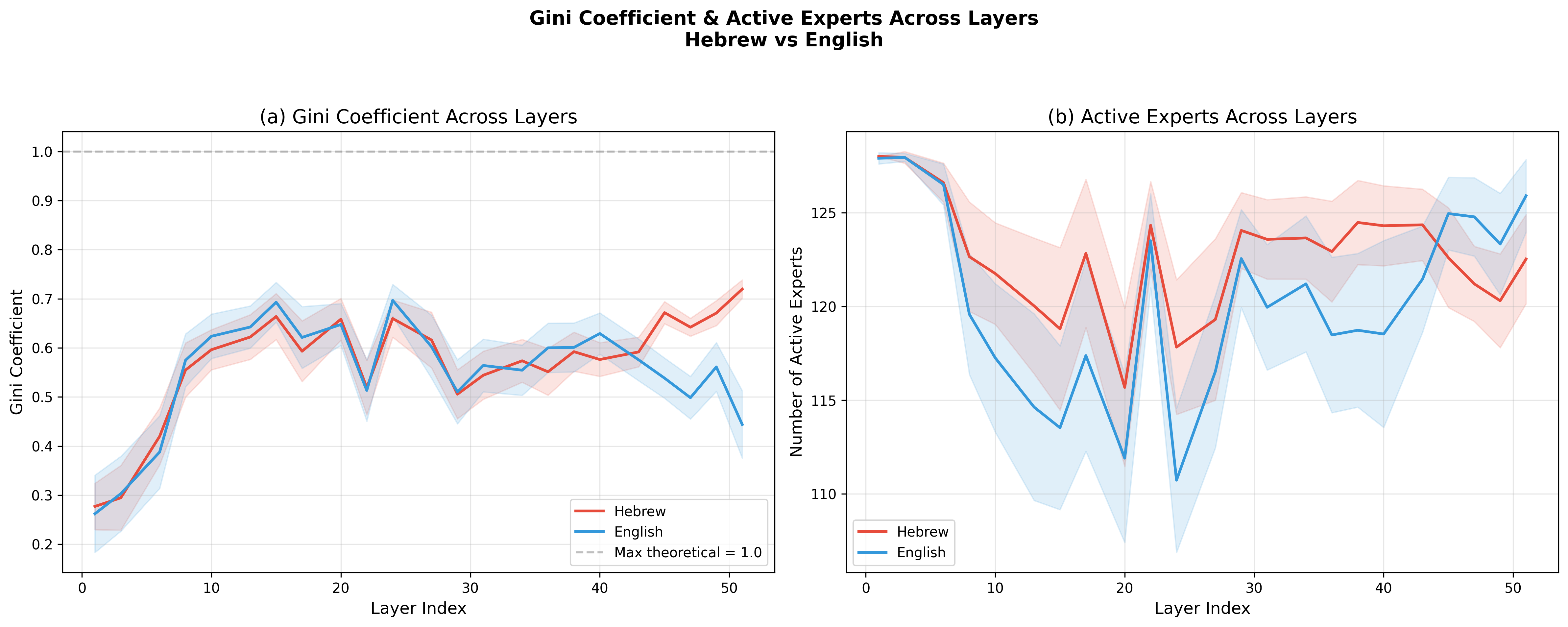}
    \caption{Nemotron base}
\end{subfigure}
\hfill
\begin{subfigure}[b]{0.48\textwidth}
    \includegraphics[width=\textwidth]{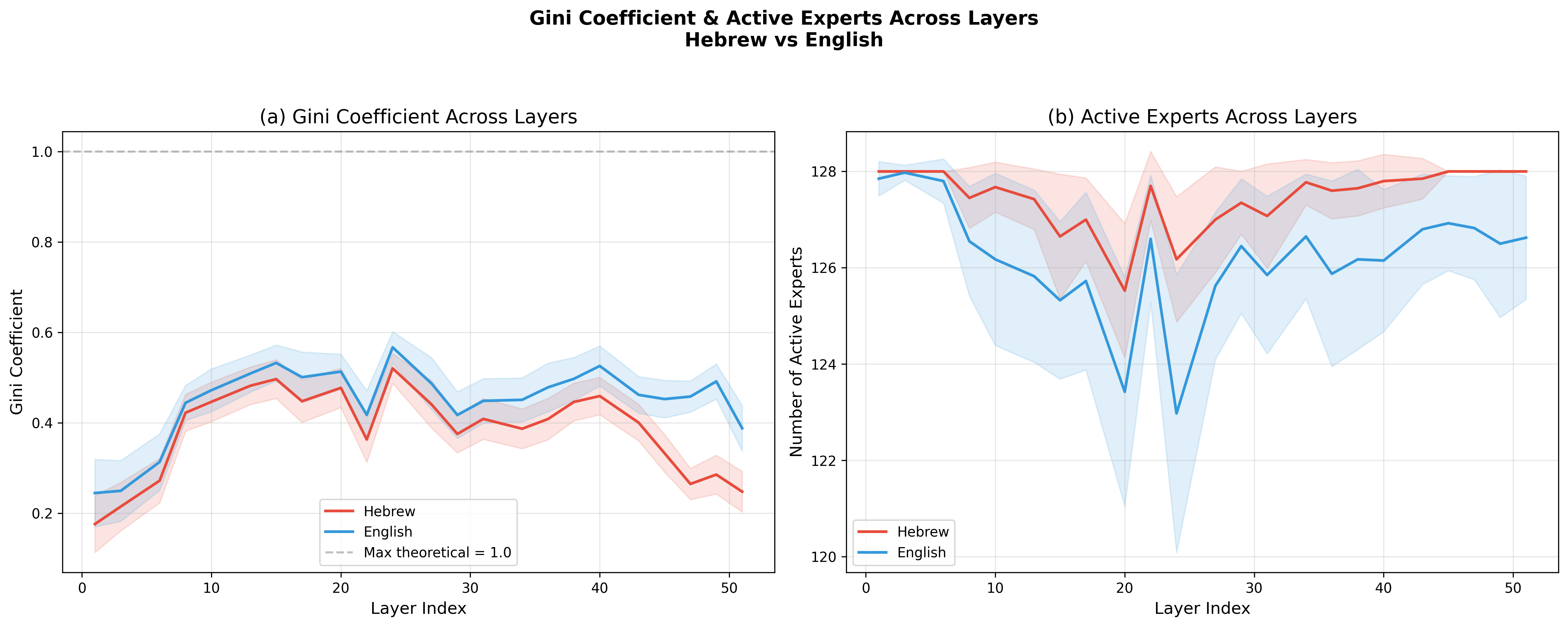}
    \caption{Nemotron base256}
\end{subfigure}
\hfill
\begin{subfigure}[b]{0.48\textwidth}
    \includegraphics[width=\textwidth]{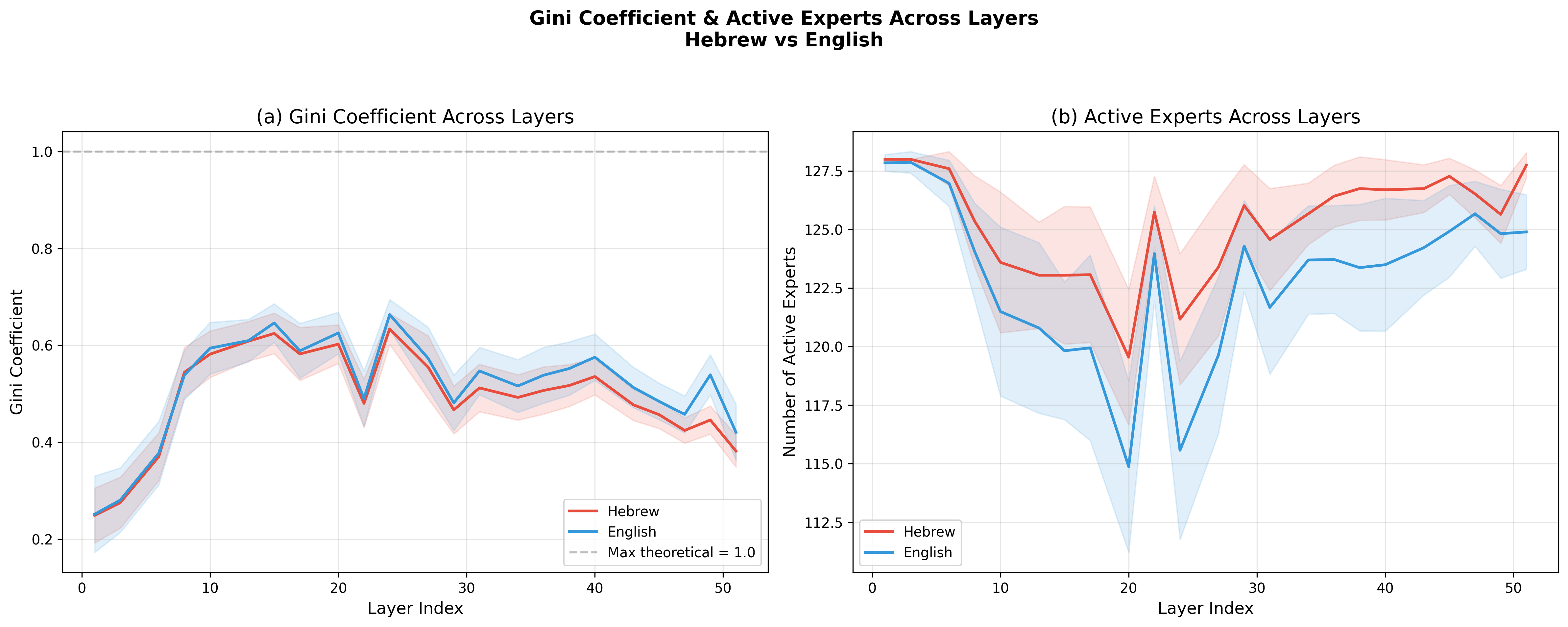}
    \caption{Nemotron instruct64}
\end{subfigure}
\caption{Gini coefficient and active expert count across MoE layers for all model variants. Each panel displays (a) the layer-wise Gini coefficient for Hebrew (red) and English (blue), (b) the per-layer Gini gap (English $-$ Hebrew), and (c) the number of active experts per layer. Panels correspond to: (a) Qwen3 base; (b) Qwen3 CPT; (c) Nemotron base; (d) Nemotron base256; (e) Nemotron instruct64.}
\label{fig:gini}
\end{figure}

\subsection{Language Specialization of Experts}
\label{sec:specialization}

To examine whether individual experts develop language-specific roles, we computed the Language Specificity Index (LSI) for each expert, averaged across all MoE layers. The LSI measures, for each expert, the relative difference between its mean routing weight when processing Hebrew tokens versus English tokens: values above $+0.1$ indicate Hebrew-specialized experts, values below $-0.1$ indicate English-specialized experts, and values within the $\pm 0.1$ neutral zone indicate shared (language-agnostic) experts. Figure~\ref{fig:specialization} presents the per-expert LSI profiles (panel a), the distribution of LSI values across all experts (panel b), and the resulting categorization.

\subsubsection{Base Models}
\label{sec:specialization_base}

The two base models reveal markedly different degrees of language specialization prior to any fine-tuning.

The \textbf{Nemotron base model} (Figure~\ref{fig:specialization}b) exhibits moderate language-specific routing: out of 128 experts, 18 are classified as Hebrew-specific (14.1\%), 16 as English-specific (12.5\%), and 94 as shared (73.4\%). The language-specific experts are roughly balanced between the two languages, and the majority of experts fall within the neutral zone. The LSI distribution (panel b) is centered near zero with moderate tails, and the per-expert profile (panel a) shows a mix of clearly language-specific bars (red and blue) alongside a large population of grey shared experts.

The \textbf{Qwen base model} (Figure~\ref{fig:specialization}a) exhibits a substantially higher degree of language specialization. Of the 128 experts, 28 are classified as Hebrew-specific (21.9\%), 44 as English-specific (34.4\%), and only 56 as shared (43.8\%). The proportion of shared experts is dramatically lower than in the Nemotron base model (43.8\% vs.\ 73.4\%), indicating that the Qwen architecture develops stronger language-specific routing at baseline. The LSI distribution is broader, with a larger number of experts exhibiting high absolute LSI values, and the per-expert profile (panel a) shows more pronounced and more numerous red and blue bars extending well beyond the $\pm 0.1$ thresholds. More than half of all experts (56.2\%) display measurable language preference, compared to only approximately one quarter in the Nemotron model. The distribution is also notably asymmetric, with a larger number of English-specific experts (44) than Hebrew-specific experts (28), suggesting that the expert pool is likely disproportionately tuned toward English processing.

\subsubsection{Effect of Fine-Tuning}
\label{sec:specialization_ft}

Fine-tuning on the Hebrew-enriched corpus produces a substantial shift toward language-agnostic routing in both architectures.

\begin{figure}[H]
\centering
\begin{subfigure}[b]{0.48\textwidth}
    \includegraphics[width=\textwidth]{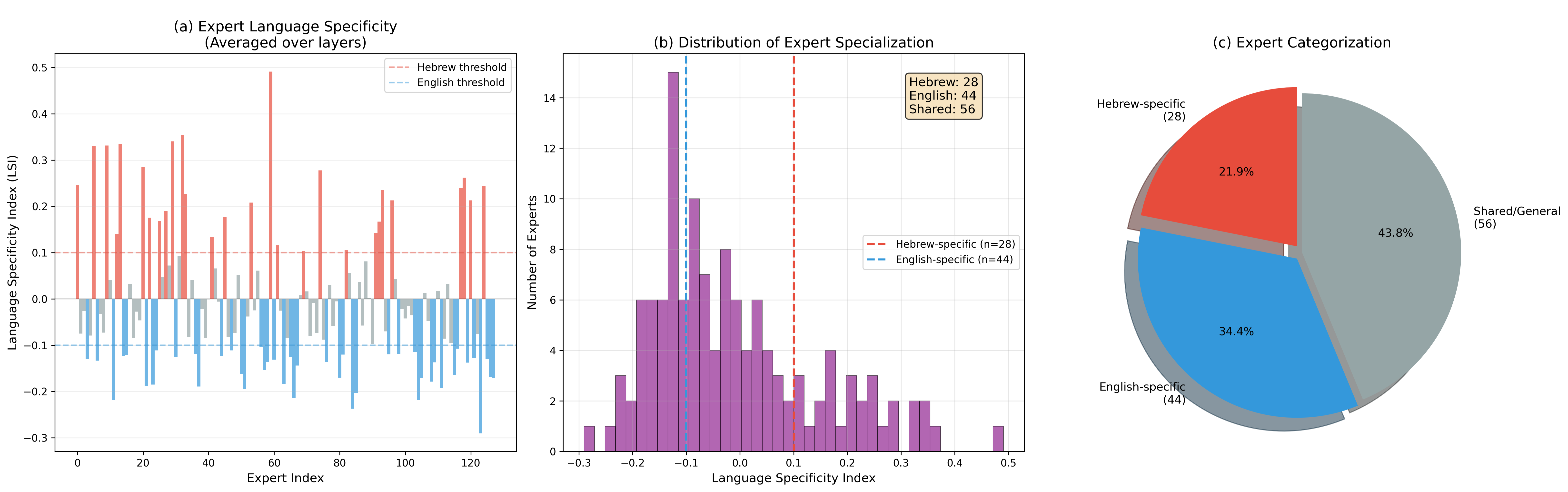}
    \caption{Qwen3 base}
\end{subfigure}
\hfill
\begin{subfigure}[b]{0.48\textwidth}
    \includegraphics[width=\textwidth]{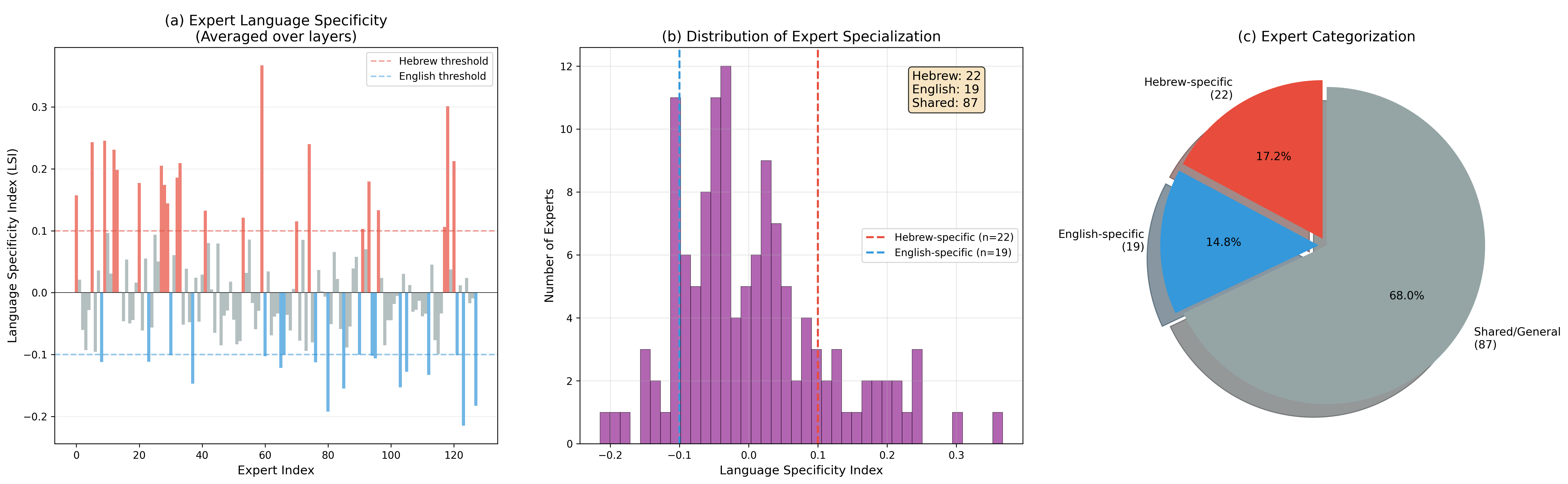}
    \caption{Qwen3 CPT}
\end{subfigure}
\\[0.5em]
\begin{subfigure}[b]{0.48\textwidth}
    \includegraphics[width=\textwidth]{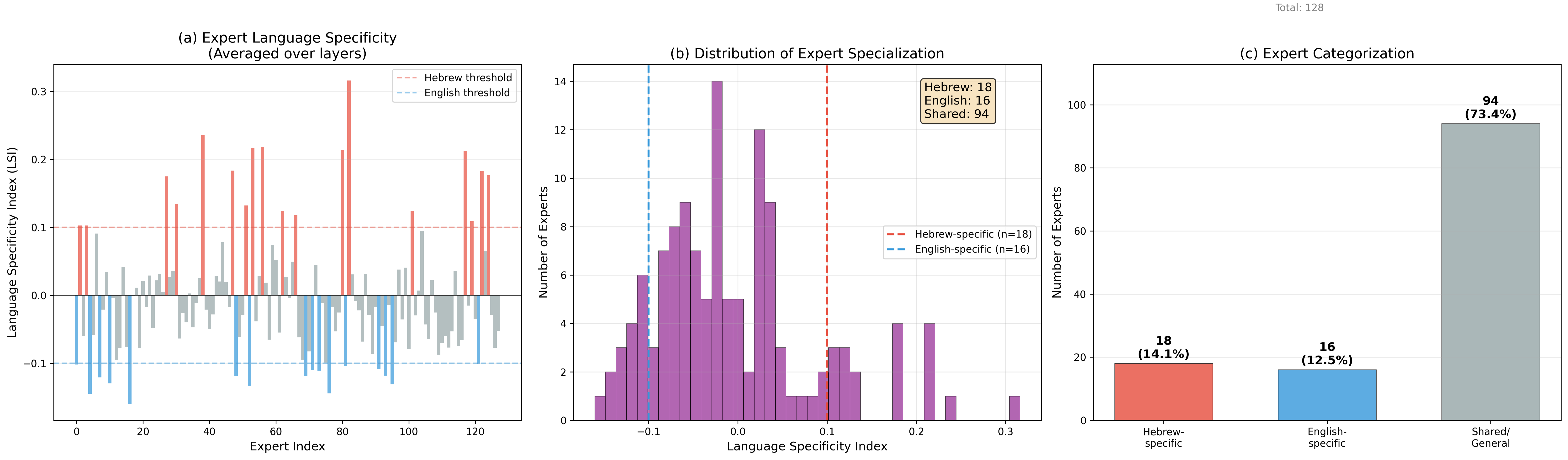}
    \caption{Nemotron base}
\end{subfigure}
\hfill
\begin{subfigure}[b]{0.48\textwidth}
    \includegraphics[width=\textwidth]{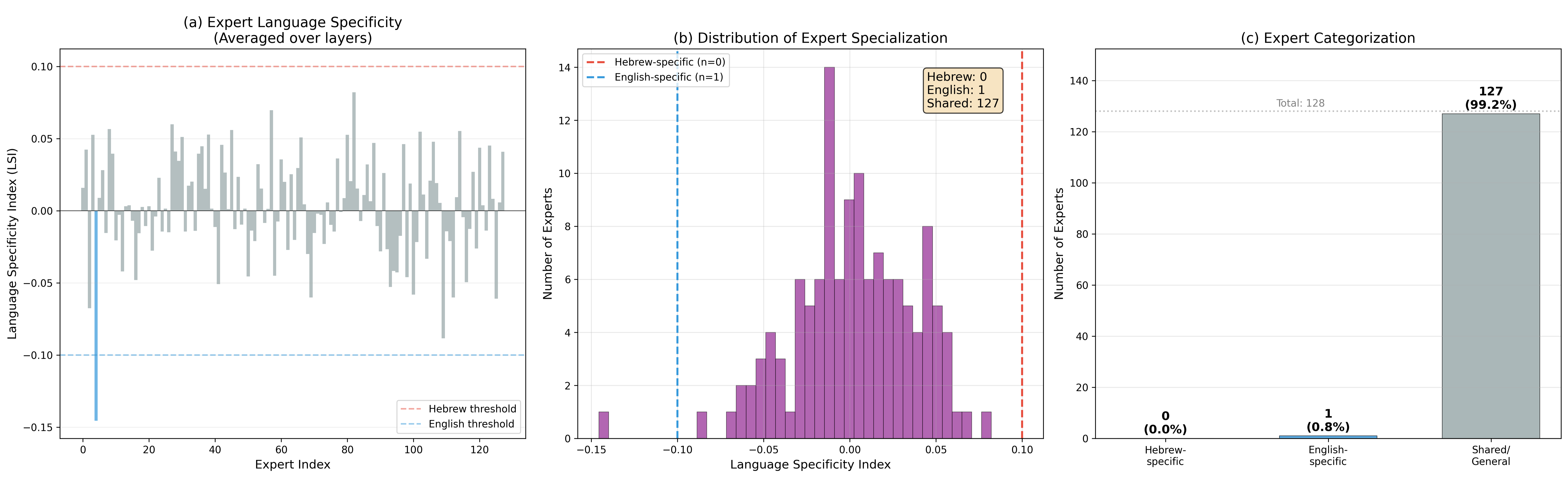}
    \caption{Nemotron base256}
\end{subfigure}
\\[0.5em]
\begin{subfigure}[b]{0.48\textwidth}
    \includegraphics[width=\textwidth]{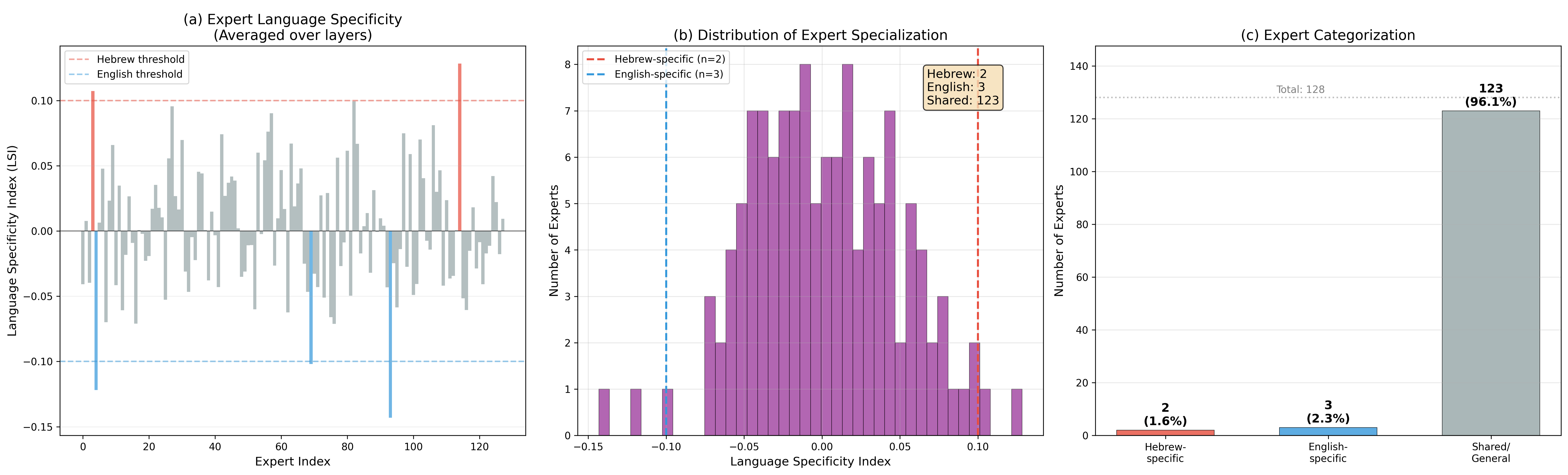}
    \caption{Nemotron instruct64}
\end{subfigure}
\caption{Expert language specialization analysis for all model variants. Each panel displays (left) the per-expert Language Specificity Index (LSI), averaged across all MoE layers, with Hebrew-specific experts shown in red and English-specific experts in blue; (center) the histogram of LSI values across all 128 experts, with dashed lines indicating the $\pm 0.1$ classification thresholds; and (right) the resulting categorization of experts into Hebrew-specific, English-specific, and shared categories. Panels correspond to: (a) Qwen3 base; (b) Qwen3 CPT; (c) Nemotron base; (d) Nemotron base256; (e) Nemotron instruct64.}
\label{fig:specialization}
\end{figure}

In the \textbf{CPT Qwen model} (Figure~\ref{fig:specialization}b), the number of language-specific experts decreases relative to the base model: 22 experts are classified as Hebrew-specific (17.2\%), 19 as English-specific (14.8\%), and 87 as shared (68.0\%). The LSI distribution (panel b) narrows compared to the base model, with more experts concentrated within or near the $\pm 0.1$ neutral zone. While a minority of experts retain language-specific routing, the overall trend is a clear shift toward shared utilization.

The Nemotron model exhibits an even more pronounced version of this effect under both training strategies. In the \textbf{Nemotron instruct64 model} (Figure~\ref{fig:specialization}e), language specialization is further reduced: only 2 experts are classified as Hebrew-specific (1.6\%), 3 as English-specific (2.3\%), and 123 as shared (96.1\%). The LSI distribution narrows substantially, with nearly all experts falling within the neutral zone.

The most extreme case is the \textbf{Nemotron base256 model} (Figure~\ref{fig:specialization}d), in which the training pipeline has driven expert specialization to near-complete language agnosticism: 0 experts are classified as Hebrew-specific, only 1 as English-specific (0.8\%), and 127 as shared (99.2\%). The LSI values are tightly clustered near zero, forming a narrow unimodal distribution. With only a single expert classified as language-specific out of 128, this model suggests that CPT \citep{gururangan2020} can substantially reorganize the routing structure of an MoE model, virtually eliminating language-specific pathways.

\subsubsection{Summary}
\label{sec:specialization_summary}

Across all model variants, a consistent pattern emerges: fine-tuning on balanced bilingual data reduces language-specific expert specialization and shifts the routing structure toward shared, language-agnostic experts. This effect is present in both architectures but is most pronounced in the Nemotron model, particularly under the base256 training pipeline, where language-specific experts are nearly entirely eliminated. These findings suggest that after sufficient exposure to balanced bilingual data, experts may develop specializations along dimensions other than language identity, such as syntactic structures, semantic domains, or discourse functions, that generalize across languages, rather than maintaining dedicated language-specific processing pathways.

\subsection{Cross-Lingual Activation Similarity}
\label{sec:cosine_similarity}

\begin{figure}[!t]
\centering
\begin{subfigure}[b]{0.48\textwidth}
    \includegraphics[width=\textwidth]{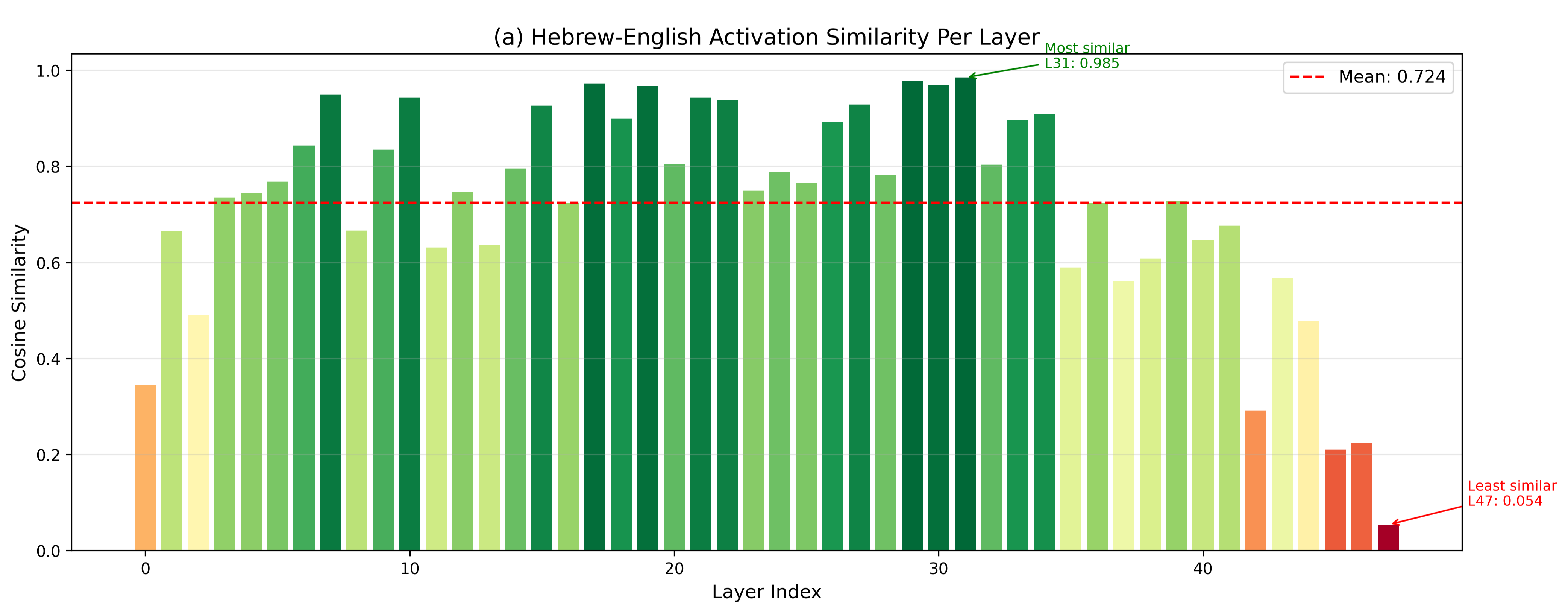}
    \caption{Qwen3 base}
\end{subfigure}
\hfill
\begin{subfigure}[b]{0.48\textwidth}
    \includegraphics[width=\textwidth]{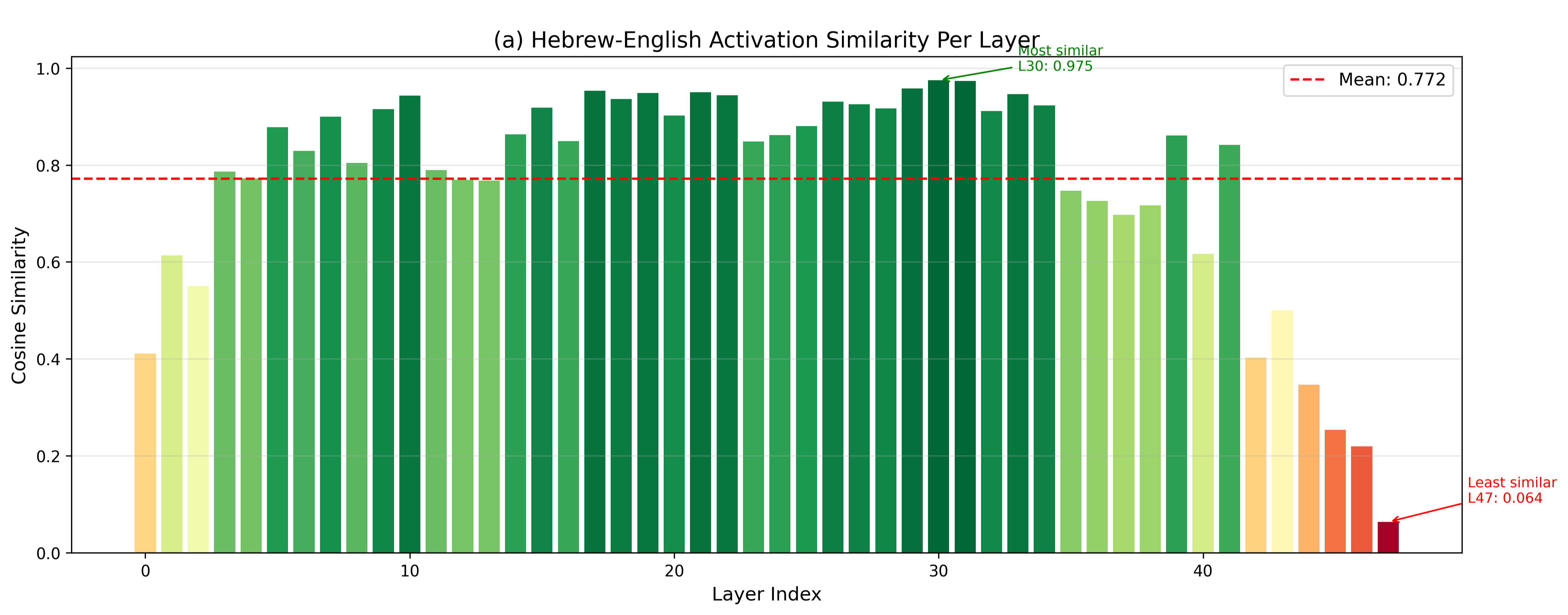}
    \caption{Qwen3 CPT}
\end{subfigure}

\vspace{1.5em} 

\begin{subfigure}[b]{0.48\textwidth}
    \includegraphics[width=\textwidth]{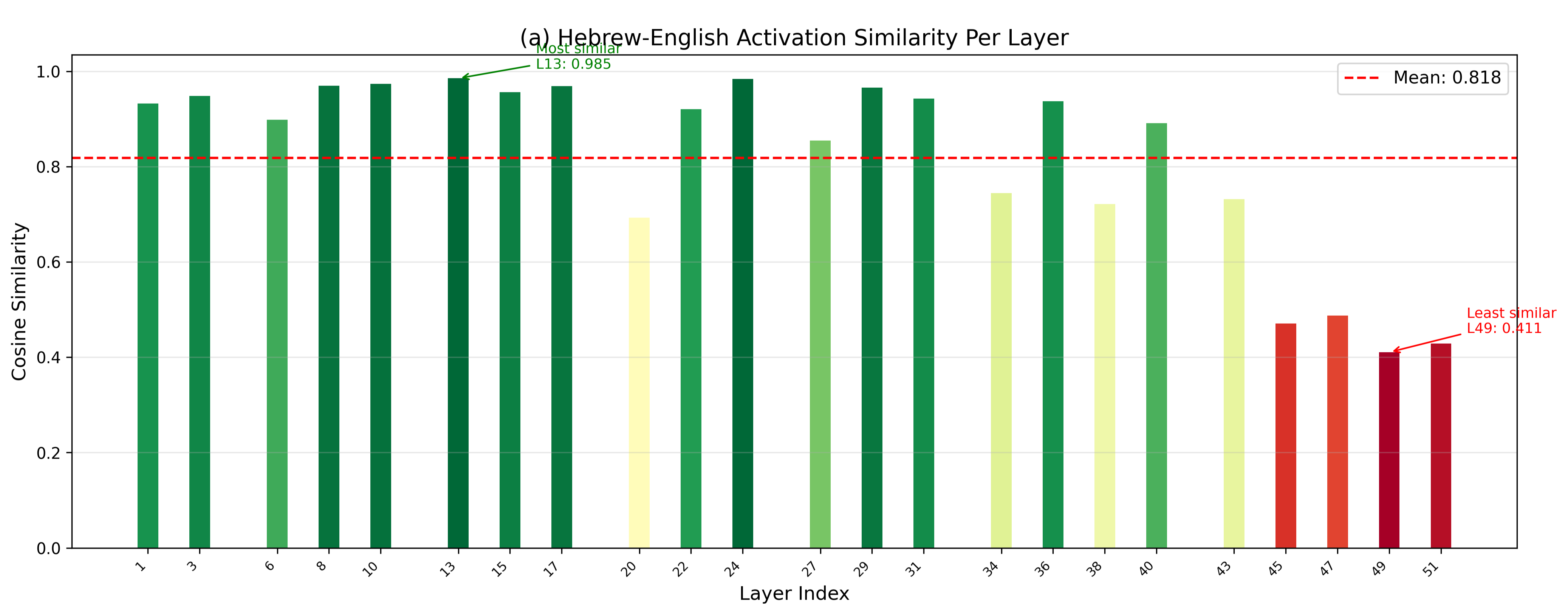}
    \caption{Nemotron base}
\end{subfigure}
\hfill
\begin{subfigure}[b]{0.48\textwidth}
    \includegraphics[width=\textwidth]{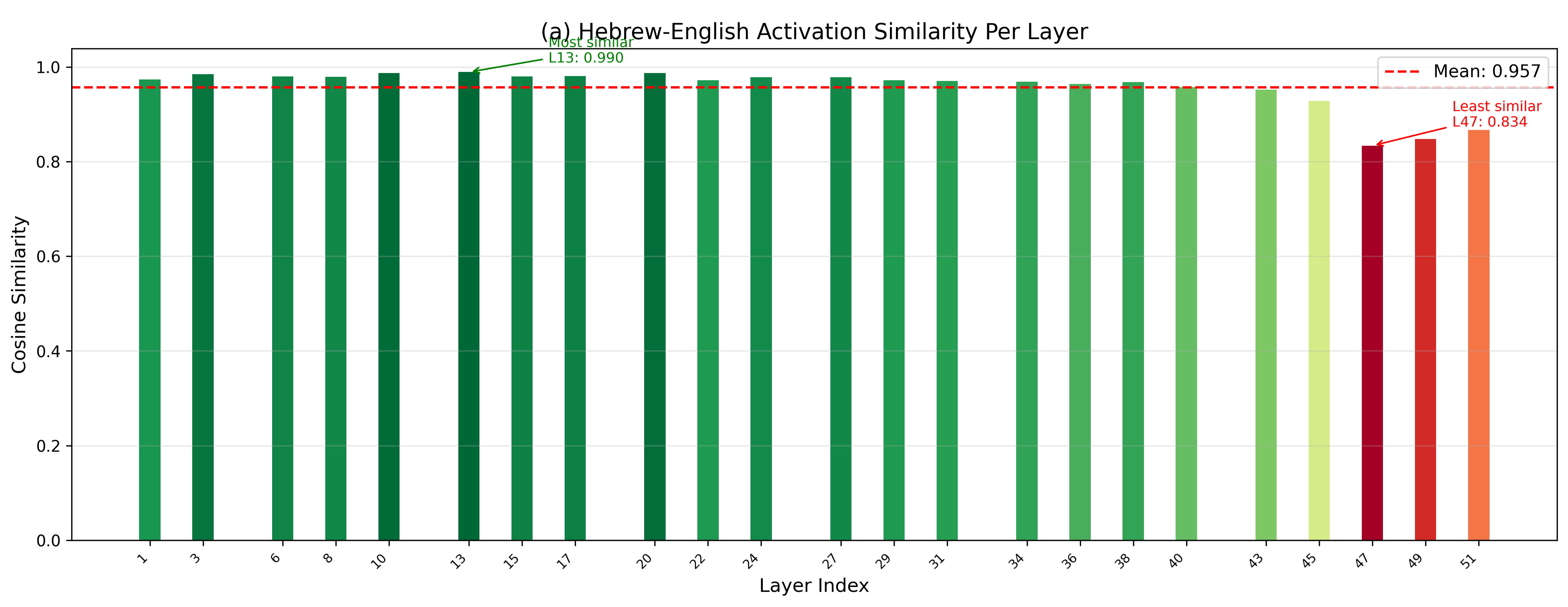}
    \caption{Nemotron base256}
\end{subfigure}

\vspace{1.5em}

\begin{subfigure}[b]{0.48\textwidth}
    \centering
    \includegraphics[width=\textwidth]{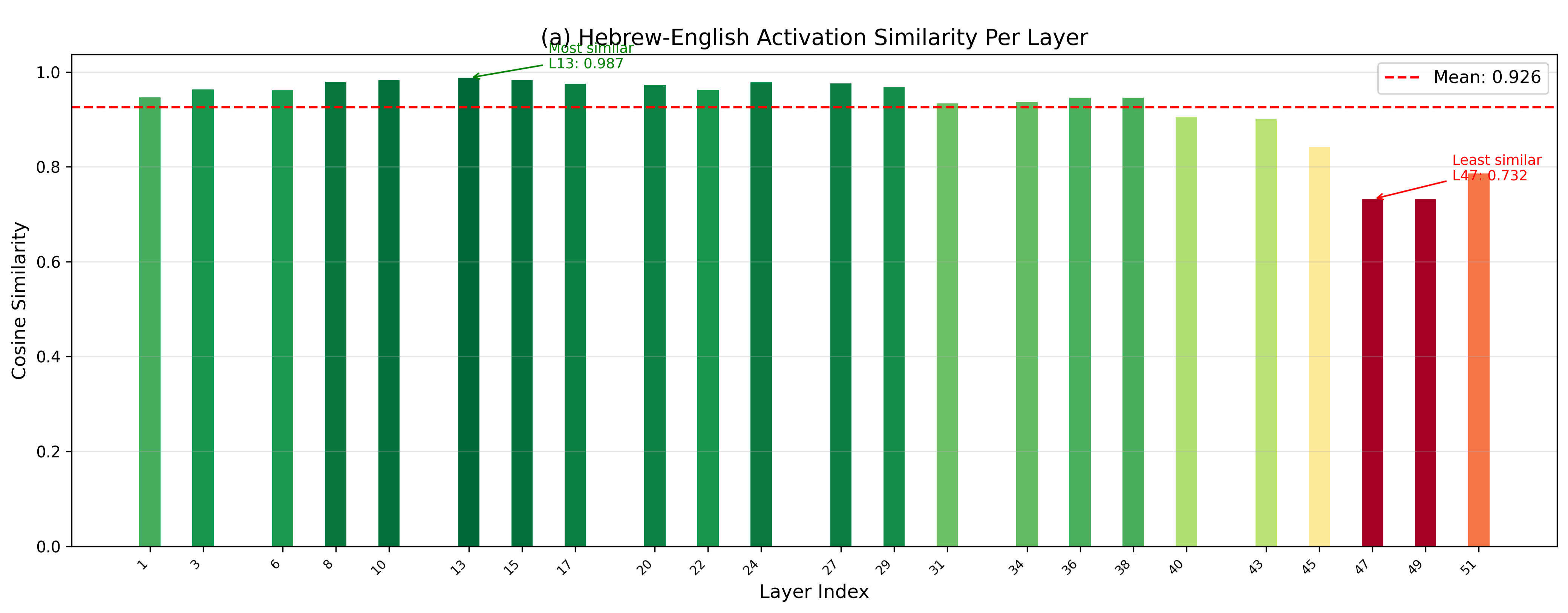}
    \caption{Nemotron instruct64}
\end{subfigure}

\caption{Cross-lingual activation similarity per layer for all model variants. Each panel displays the per-layer cosine similarity between Hebrew and English expert activation vectors, with bars color-coded from green (high similarity) to red (low similarity). The red dashed line indicates the mean cosine similarity across all layers. Annotations highlight the most similar and least similar layers. Panels correspond to: (a)~Qwen3 base; (b)~Qwen3 CPT; (c)~Nemotron base; (d)~Nemotron base256; (e)~Nemotron instruct64.}
\label{fig:cosine_similarity}
\end{figure}

Figure~\ref{fig:cosine_similarity} presents the layer-wise cosine similarity between the Hebrew and English expert utilization vectors, that is, how similarly each layer distributes tokens across experts when processing the two languages, where values close to 1.0 indicate near-identical routing patterns and values close to 0 indicate fully divergent expert utilization.

Across all model variants, the lowest cross-lingual similarity is consistently found in the deepest layers, confirming that the final layers of the network are where the greatest language-specific divergence in expert routing occurs. This is consistent with the entropy and Gini analyses reported above, and aligns with the general understanding that deeper layers perform more language-specific computations closer to the output \citep{bandarkar2026}.

The two Qwen models (base and CPT) additionally exhibit reduced similarity in the earliest layers, producing a characteristic U-shaped profile in which the middle layers show the highest cross-lingual similarity while both the input-proximal and output-proximal layers diverge. This pattern is consistent with the view that boundary layers play a dual role: early layers must handle language-specific surface features (script, tokenization patterns, morphological cues), while deep layers handle language-specific abstract representations closer to the output \citep{tenney2019, belinkov2017evaluating}.

The Qwen base model shows lower overall similarity (mean = 0.724) compared to the Nemotron base model (mean = 0.818), consistent with the stronger language specialization documented in Section~\ref{sec:specialization}. Fine-tuning raises the mean similarity across all models, with the most pronounced improvement in the Nemotron variants, the base256 model increases from 0.818 to 0.957 and the instruct64 to 0.926, indicating that bilingual fine-tuning drives the two languages toward more convergent routing patterns.

\subsection{Cross-Linguistic Generalization: Japanese}
\label{sec:japanese}

To assess whether deep-layer routing collapse is specific to Hebrew or 
reflects a more general property of MoE systems processing pre-training underrepresented 
languages, we extended the routing analysis to Japanese. Japanese is 
typologically distant from Hebrew in every major dimension: it is 
agglutinative and head-final, written in three interlaced scripts (hiragana, 
katakana, and kanji), and lacks whitespace delimiters between words. Yet despite these typological distances, Japanese shares with Hebrew a similar degree of underrepresentation relative to dominant pre-training languages such as English and Chinese in large-scale pre-training corpora. Three model variants sharing the 
identical Qwen3-30B-A3B base \citep{alibaba2025qwen3} architecture were analyzed: the original 
pre-trained base model; Qwen3-Swallow-30B-A3B-CPT \citep{fujii2024continual}, which underwent CPT on a large Japanese corpus, and Shisa-Qwen3-30B \citep{shisa2026checkpoint079}, which was fine-tuned directly from the 
base checkpoint using SFT without any intermediate CPT stage. Because all three models share the same architecture, differences in routing behavior are less confounded by structural variance, though they may still reflect disparities in data composition, optimization recipes, and checkpoint histories. To preserve thematic consistency across evaluations, all Japanese routing measurements were conducted on content translated into Japanese from the exact same \texttt{ted\_he\_en\_chunks} and \textbf{FLORES-200} benchmark \citep{nllb2022} configurations used throughout this work.

\subsubsection{Deep-Layer Routing Collapse in the Base Model}
\label{sec:japanese_base}

The Qwen3-30B-A3B-Base model \citep{alibaba2025qwen3} exhibits a routing collapse for Japanese that 
is qualitatively similar to the Hebrew collapse documented in 
Section~\ref{sec:usage_collapse}. Entropy remains broadly stable through 
the middle layers then drops sharply in the final layers, closely mirroring 
the Hebrew base model profile (Figure~\ref{fig:japanese}a). The Gini 
coefficient rises monotonically across layers and spikes sharply in the 
deep layers, again matching the Hebrew pattern (Figure~\ref{fig:japanese}c). 
The expert specialization structure further parallels the Hebrew case: 
of the 128 experts, 25 are Japanese-specific (19.5\%), 36 are 
English-specific (28.1\%), and 67 are shared (52.3\%) 
(Figure~\ref{fig:japanese}e, panel C), a distribution nearly identical 
to the Hebrew base model (21.9\% Hebrew-specific, 34.4\% English-specific, 
43.8\% shared, panel A). The similar collapse signatures across two 
typologically unrelated languages are consistent with pre-training 
underrepresentation as a contributing factor rather 
than any language-specific property.

\subsubsection{Effect of Training Strategy}
\label{sec:japanese_ft}

\paragraph{CPT (Qwen3-Swallow-30B-A3B-CPT).}
The CPT pipeline substantially corrects deep-layer collapse across 
both metrics. Entropy in the Swallow model remains stable through the deep 
layers rather than collapsing (Figure~\ref{fig:japanese}a), and the Gini 
spike is substantially attenuated relative to the base model 
(Figure~\ref{fig:japanese}c). Expert specialization shifts toward language 
agnosticism: 23 experts are Japanese-specific (18.0\%), 20 are 
English-specific (15.6\%), and 85 are shared (66.4\%) 
(Figure~\ref{fig:japanese}e, panel D), a pattern qualitatively 
consistent with the Hebrew CPT result of 68.0\% shared experts 
(panel B), though we note this comparison spans different languages and 
training corpora and should be interpreted with caution.

\paragraph{SFT-only (Shisa-Qwen3-30B).}
The SFT-only model improves modestly over the base in the middle layers 
but converges almost completely with the base model in the deep layers, with both 
reaching the same terminal entropy and Gini values 
(Figure~\ref{fig:japanese}b, d). This indicates that, in this comparison, SFT without prior CPT provides limited remediation of deep-layer routing structure.

\begin{figure}[H]
\centering
\begin{subfigure}[b]{0.48\textwidth}
    \includegraphics[width=\textwidth]{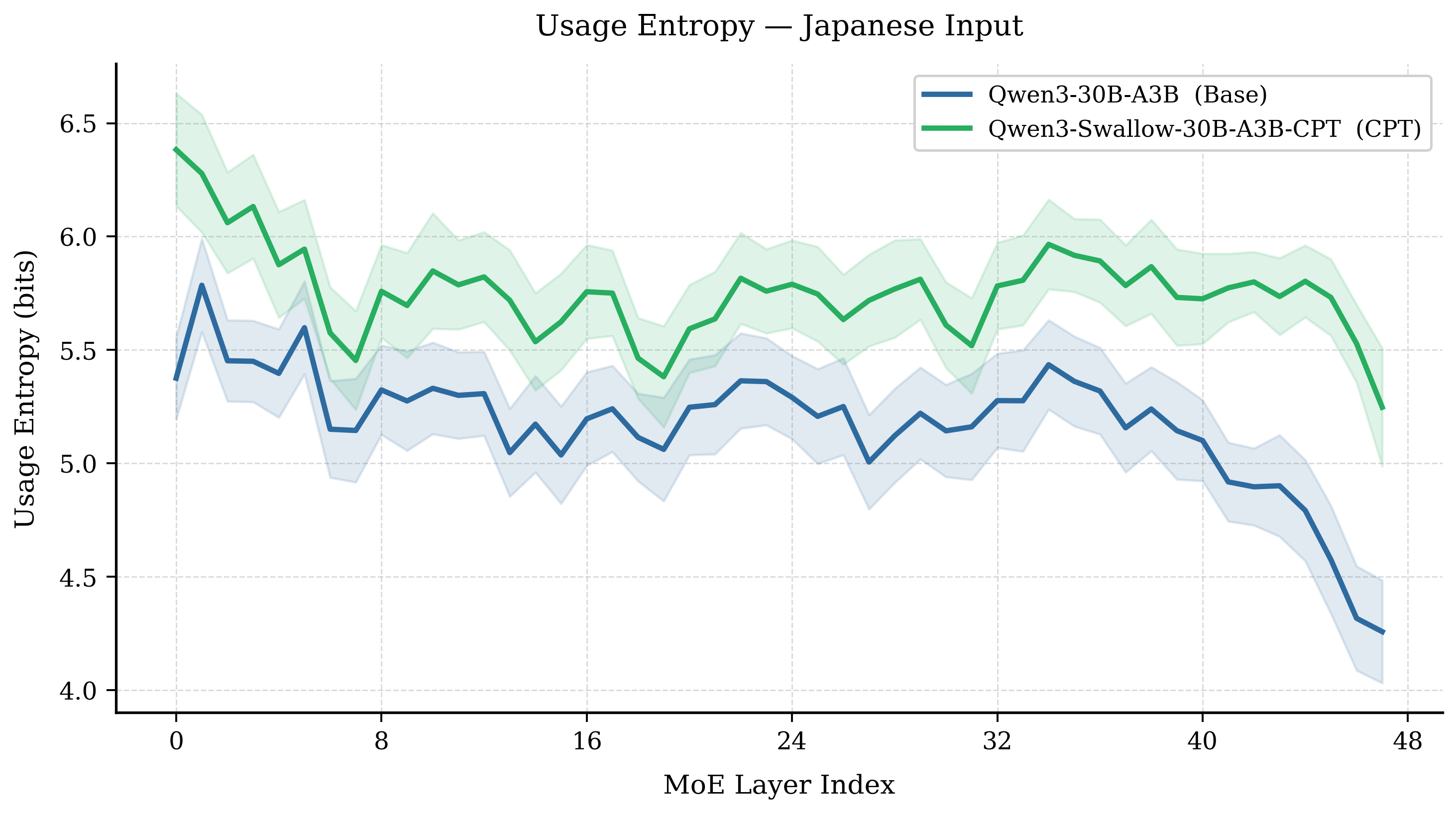}
    \caption{Entropy: Base vs. CPT (Swallow)}
\end{subfigure}
\hfill
\begin{subfigure}[b]{0.48\textwidth}
    \includegraphics[width=\textwidth]{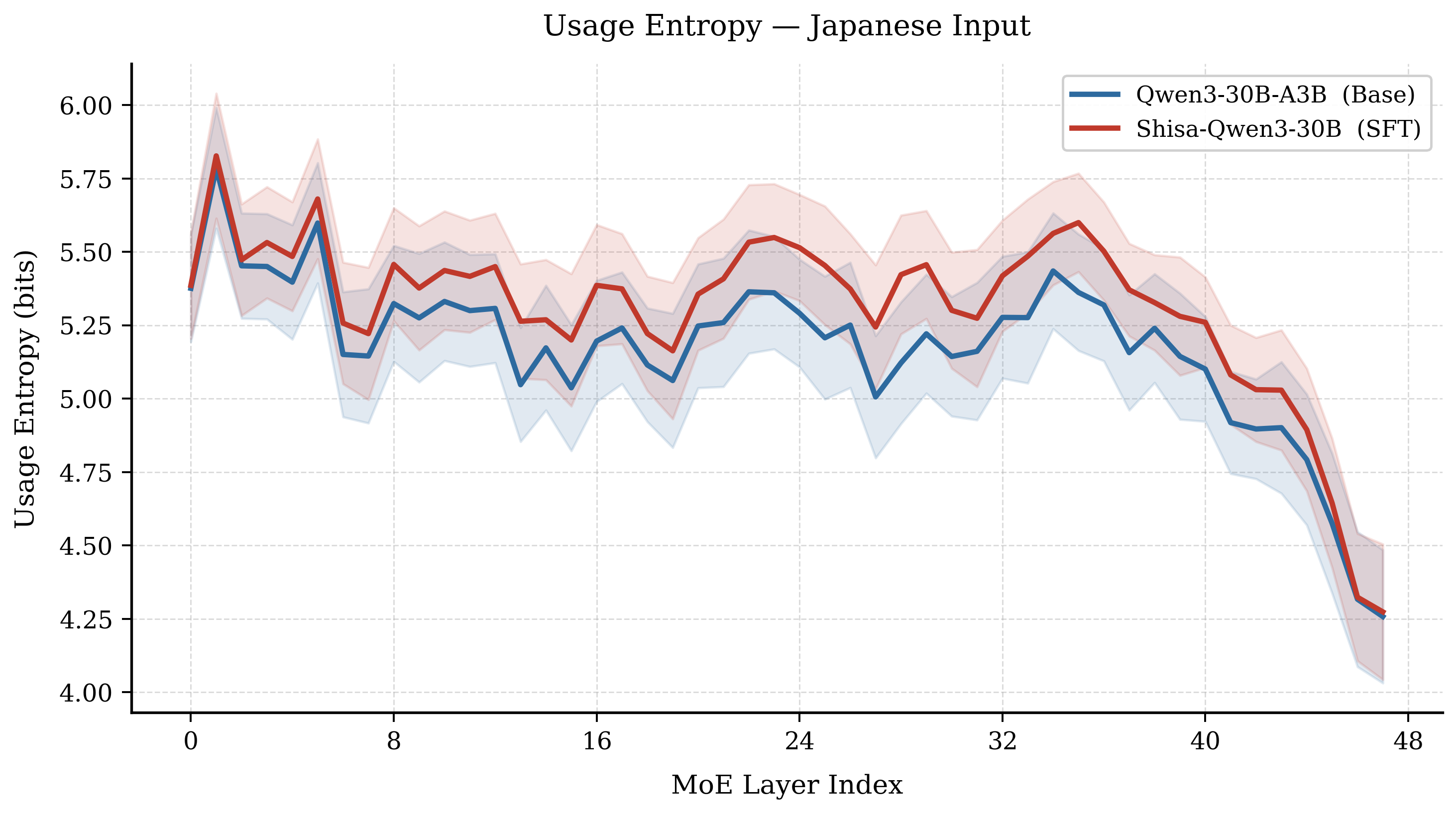}
    \caption{Entropy: Base vs. SFT (Shisa)}
\end{subfigure}

\vspace{0.8em} 

\begin{subfigure}[b]{0.48\textwidth}
    \includegraphics[width=\textwidth]{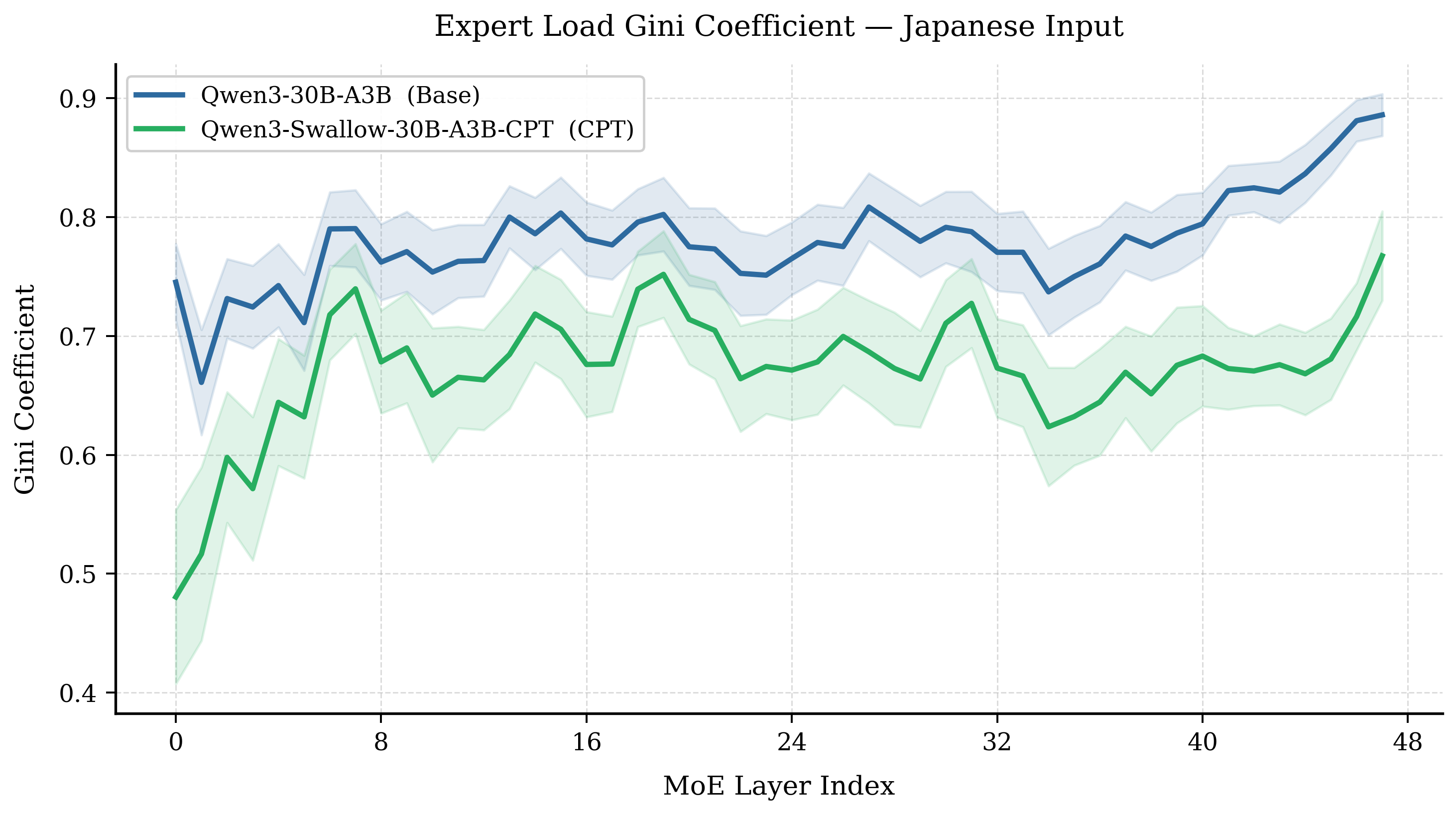}
    \caption{Gini: Base vs. CPT (Swallow)}
\end{subfigure}
\hfill
\begin{subfigure}[b]{0.48\textwidth}
    \includegraphics[width=\textwidth]{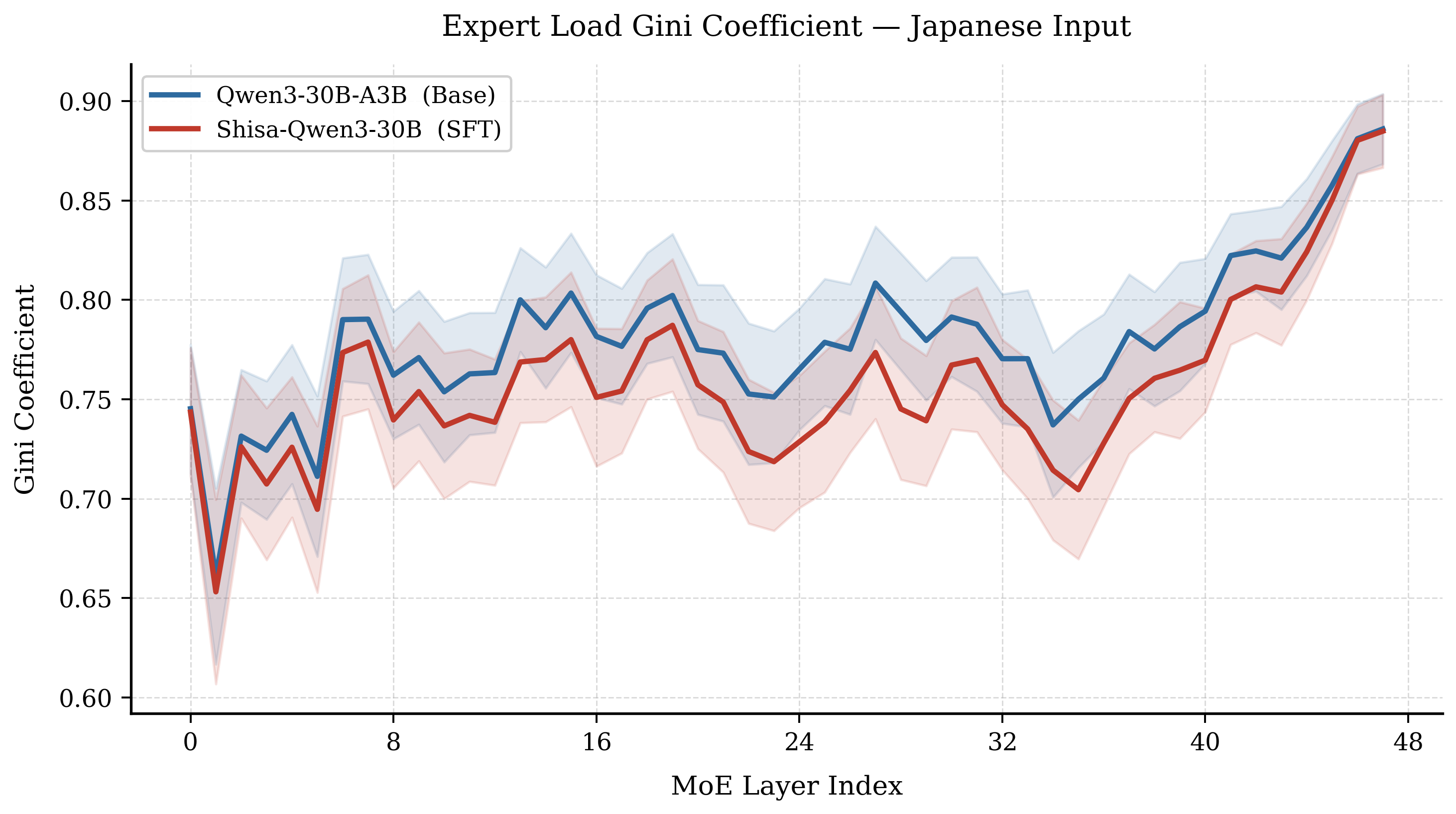}
    \caption{Gini: Base vs. SFT (Shisa)}
\end{subfigure}

\vspace{0.8em}

\begin{subfigure}[b]{0.9\textwidth}
    \centering
    \includegraphics[width=\textwidth]{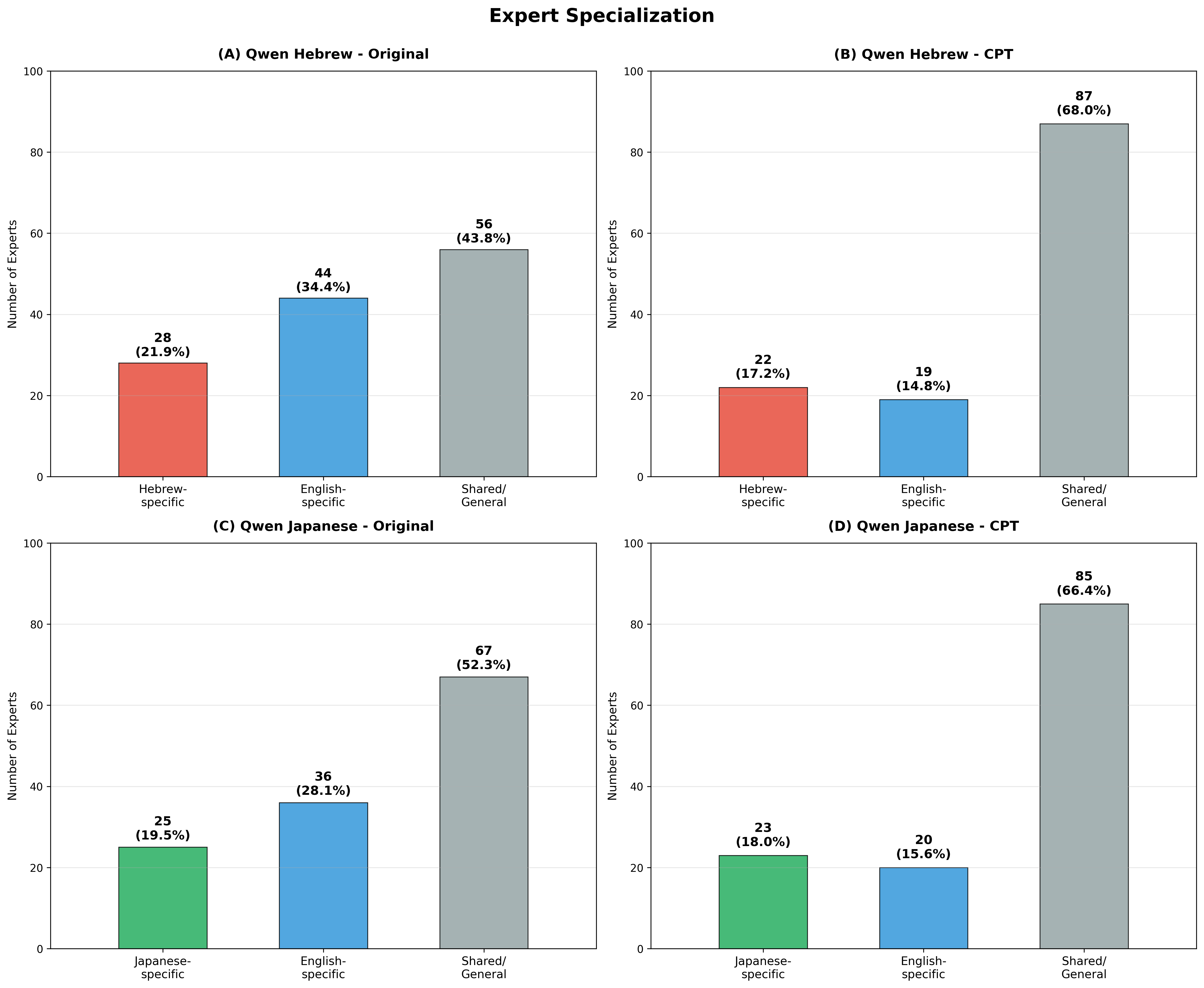}
    \caption{Expert specialization: Hebrew and Japanese compared}
\end{subfigure}
\end{figure}
\begin{figure}[t]
\ContinuedFloat
\caption{Japanese routing analysis across three model variants sharing the identical Qwen3-30B-A3B base architecture. Panels (a) and (b) show layer-wise routing entropy with shaded $\pm$1 standard deviation bands. Panels (c) and (d) show the corresponding Gini coefficient profiles. Panel (e) shows expert specialization counts for Hebrew and Japanese under base and CPT conditions: panels A and C show the base model distributions for Hebrew and Japanese respectively; panels B and D show the corresponding CPT distributions. CPT substantially corrects deep-layer collapse across all metrics while SFT shows limited remediation in the deepest layers.}
\label{fig:japanese}
\end{figure}

\subsubsection{Summary}
\label{sec:japanese_summary}

Deep-layer routing collapse is observed in Japanese with a similar characteristic profile. Entropy, Gini, and expert-specialization profiles mirror the Hebrew pattern, supporting the view that this bottleneck may generalize across typologically unrelated pre-training underrepresented languages and is not an artifact of any specific linguistic structure. CPT reliably corrects this collapse across all metrics, while SFT alone does not improve expert distribution to the same extent as models that underwent CPT. These results extend the main findings of this work beyond the Hebrew–English setting and support the interpretation that routing imbalance is a systematic consequence of pre-training data representation levels \citep{lauscher2020} rather than language-intrinsic properties.

\subsection{Routing Dynamics and Downstream Performance}
\label{sec:benchmarks}
\subsubsection{Hebrew}
To assess whether the routing improvements documented in Sections~\ref{sec:usage_entropy}--\ref{sec:japanese} are associated with measurable gains in downstream task performance, we evaluate base and CPT models on the Hebrew LLM Leaderboard \citep{shmidman2024dictalm2} using the leaderboard's few-shot evaluation protocol, as these models are not designed to follow chat-style prompts directly. Table~\ref{tab:hebrew_base} reports results for the base and CPT models across six tasks: SNLI, QA (TLNLS), Sentiment, Winograd, Translation (BLEU), and Israeli Trivia. Table~\ref{tab:hebrew_instruct} reports results for the instruct-tuned variants on a complementary Hebrew benchmark set \citep{weinberger2026hebatron}, evaluated in a zero-shot setting, as SFT models are designed to follow prompts directly without requiring few-shot demonstrations.
 
For the CPT models, routing improvement and performance gains align closely. Nemotron base256, which largely eliminates deep-layer routing collapse (Section~\ref{sec:usage_nemotron_ft}), achieves the strongest overall gains, with a Hebrew average rising from 65.61 to 68.0 (+2.39 points). The improvement is especially pronounced on Israeli Trivia (58.14 $\to$ 72.1) — a task that plausibly reflects Hebrew-specific cultural and world knowledge — consistent with the routing analysis showing that CPT drives expert utilization toward broader, more language-agnostic patterns. The Qwen CPT model shows a correspondingly modest improvement (61.85 → 62.80); in our view, while the model initially improves routing efficiency, these gains are partly limited by residual routing collapse in the deepest layers (Section~\ref{sec:usage_qwen_ft}). 
 
For the SFT variants, the results exhibit a nuanced performance profile across different task categories. Compared to the original Nemotron-3-Nano-30B (instruct) checkpoint, the instruct64 model demonstrates clear improvements on reasoning-intensive tasks, with GSM8K  \citep{cobbe2021training, openai2024gpt4o} Hebrew rising from 60.0 to 72.0 and the Psychometric Entrance Test (Psi) increasing from 42.0 to 53.1. MMLU \citep{hendrycks2020measuring} (HE) also shows a modest gain from 60.0 to 62.0. However, this specialization corresponds with selective trade-offs on individual knowledge-based benchmarks: Hellaswag \citep{zellers2019hellaswag} (HE) drops sharply from 70.0 to 52.0, and ARC \citep{clark2018think} (HE) declines slightly from 90.0 to 89.0. While the overall Hebrew average still improves modestly (64.4 → 65.6), the within-task volatility — large gains on reasoning paired with large drops on commonsense — reflects a model that significantly improves on certain tasks while losing capabilities it previously possessed. This pattern is consistent with the routing analysis, which shows that SFT alone improves expert distribution less substantially than SFT applied after CPT (Section~\ref{sec:usage_nemotron_ft}). Taken together, these results indicate that CPT-induced routing reorganization produces more consistent and comprehensive downstream gains than SFT alone, particularly for tasks requiring rich linguistic and cultural knowledge of the target language. This pattern broadly aligns with the routing analysis reported in Sections~\ref{sec:usage_entropy}--\ref{sec:specialization}: models exhibiting higher post-training usage entropy, lower Gini coefficients, and a greater proportion of shared (language-agnostic) experts tend to achieve stronger Hebrew benchmark performance, while models retaining deep-layer routing collapse and language-specific expert pathways show more limited or uneven gains (Tables~\ref{tab:hebrew_base} and~\ref{tab:hebrew_instruct}).
 
\begin{table}[H]
\centering
\caption{Hebrew benchmark results for base and CPT models. Higher is better 
for all metrics. Avg = macro-average across the six tasks.}
\label{tab:hebrew_base}
\resizebox{\textwidth}{!}{%
\begin{tabular}{llcccccccc}
\toprule
\textbf{Model} & \textbf{Training} &
\textbf{SNLI} & \textbf{QA (TLNLS)} & \textbf{Sentiment} &
\textbf{Winograd} & \textbf{Translation} & \textbf{Israeli Trivia} &
\textbf{Avg} \\
\midrule
Qwen3-30B-A3B          & Base & \textbf{69.56} & 55.53          & 68.75          & \textbf{61.00} & 55.42          & \textbf{60.97} & 61.85          \\
Qwen3-30B-A3B (CPT)    & CPT  & 62.68          & \textbf{61.02} & \textbf{71.25} & 60.50          & \textbf{62.70} & 58.07          & \textbf{62.70} \\[4pt]
Nemotron-3-Nano-30B    & Base & 88.81          & 71.09          & \textbf{68.40} & 74.10          & 33.10          & 58.14          & 65.61          \\
Nemotron-3-Nano-30B (base256) & CPT  & \textbf{91.20} & \textbf{73.20} & 63.50 & \textbf{74.10} & \textbf{33.60} & \textbf{72.10} & \textbf{68.00} \\
\bottomrule
\end{tabular}%
}
\end{table}
 
\begin{table}[H]
\centering
\caption{Hebrew benchmark results for SFT models. Psi$^\dagger$ = Psychometric Entrance Test (Hebrew sections only). Higher is better for all metrics. Avg = macro-average across the five Hebrew tasks.}
\label{tab:hebrew_instruct}
\resizebox{\textwidth}{!}{%
\begin{tabular}{llccccccc}
\toprule
\textbf{Model} & \textbf{Training} &
\textbf{MMLU (HE)} & \textbf{ARC (HE)} & \textbf{Hellaswag (He)} & \textbf{GSM8K (He)} &
\textbf{Psi$^\dagger$ (He)} &
\textbf{Avg (He)} \\
\midrule
Nemotron-3-Nano-30B (instruct) & SFT (original) & 60.00          & \textbf{90.00} & \textbf{70.00} & 60.00          & 42.00          & 64.40 \\
Nemotron-3-Nano-30B (instruct64) & SFT            & \textbf{62.00} & 89.00          & 52.00         & \textbf{72.00} & \textbf{53.10} & \textbf{65.60} \\
\bottomrule
\end{tabular}%
}
\end{table}
 
\footnotetext[1]{The Psychometric Entrance Test (Psi), produced by the National Institute
for Testing and Evaluation (NITE), is used for higher-education admissions in Israel and
evaluates verbal reasoning, quantitative reasoning, and English proficiency.
Only the localized Hebrew sections were used here.}

\subsubsection{Japanese}
\label{sec:japanese_downstream}
 
To assess whether the routing improvements documented in Section~\ref{sec:japanese}
are associated with measurable downstream gains, we evaluate all three model variants on a suite
of Japanese NLP benchmarks covering commonsense reasoning, natural language inference,
machine translation, and reading comprehension.
Table~\ref{tab:japanese_base} reports results for the two base-regime models — the
original Qwen3-30B-A3B-Base and Qwen3-Swallow-30B-A3B-CPT \citep{fujii2024continual}
— across seven tasks.
Table~\ref{tab:japanese_instruct} then extends the comparison to the SFT
variants, namely the original Qwen3-30B-A3B-Instruct checkpoint and Shisa-Qwen3-30B,
which was fine-tuned directly from the base via SFT without any intermediate CPT stage
\citep{shisa2026checkpoint079}.
 
\paragraph{CPT corrects routing collapse and improves downstream performance
(Table~\ref{tab:japanese_base}).}
Hellaswag, WMT, and JamC-QA were evaluated using the Swallow LLM Leaderboard framework \citep{swallow2025leaderboard}, while remaining tasks used the EleutherAI LM Evaluation Harness \citep{gao2023framework}.
Swallow CPT outperforms the Qwen3 base on five of seven benchmarks, with the 
largest absolute gains on JamC-QA ($+6.3$), en$\to$ja translation 
(WMT20; \citealt{barrault2020findings}, $+6.3$), Sentiment  \citep{tyqiangz2023multilingual} ($+3.0$), 
and JCommonsenseQA~\citep{kurihara2022jglue} ($+3.0$). These gains are 
consistent with the routing analysis in Section~\ref{sec:japanese}: 
CPT substantially corrects deep-layer routing collapse 
(Figures~\ref{fig:japanese}a and~\ref{fig:japanese}c), broadening expert 
utilization and driving the model toward more language-agnostic routing patterns. 
The two tasks on which the Qwen3 base retains a modest advantage, 
HellaSwag~\citep{zellers2019hellaswag} ($-4.0$) and 
Winograd~\citep{kurihara2022jglue} ($-3.3$), involve decontextualized 
commonsense prompts that may favour the broader English-dominated pre-training 
distribution rather than Japanese-specific linguistic knowledge.

Critically, this CPT-driven performance profile mirrors the Hebrew CPT result 
in Table~\ref{tab:hebrew_base}: both show the strongest gains on 
knowledge-intensive and cross-lingual tasks, consistent with a genuine broadening 
of the model's linguistic capacity, a consequence of routing reorganisation 
rather than a narrow, task-specific adaptation. Just as Israeli Trivia showed 
the largest Hebrew CPT gain, reflecting culturally grounded knowledge acquisition 
that was systematically underrepresented in the base model's pre-training corpus, 
JamC-QA and translation show the largest Japanese gains, consistent with the same 
mechanism of routing reorganisation unlocking language-specific world knowledge 
that CPT makes newly accessible to the model's expert pool.
 
\begin{table}[H]
\centering
\caption{Japanese benchmark results for Base and CPT models. Higher is better for all metrics. $Avg$ represents the macro-average across all listed tasks.}
\label{tab:japanese_base}
\resizebox{\textwidth}{!}{%
\begin{tabular}{llcccccccc}
\toprule
\textbf{Model} & \textbf{Training} &
  \textbf{Winograd} &
  \textbf{Sentiment (MLSA)} &
  \textbf{JCommonsenseQA} &
  \textbf{en-ja WMT20$^\dagger$} &
  \textbf{ja-en WMT$^\dagger$} &
  \textbf{JamC-QA$^\dagger$} &
  \textbf{HellaSwag$^\dagger$} &
  \textbf{Avg} \\
\midrule
Qwen3-30B-A3B         & Base & \textbf{79.1} & 75.0          & 92.0          & 17.7          & 21.5          & 45.5          & \textbf{88.0} & 59.83 \\
Qwen3-Swallow-30B-A3B & CPT  & 75.8          & \textbf{78.0} & \textbf{95.0} & \textbf{24.0} & \textbf{22.2} & \textbf{51.8} & 84.0          & \textbf{61.54} \\
\bottomrule
\end{tabular}%
}
\end{table}
 
\paragraph{SFT without prior CPT redistributes rather than broadens competencies
(Table~\ref{tab:japanese_instruct}).}

Table~\ref{tab:japanese_instruct} reports results for the two SFT variants.
Shisa yields clear gains on knowledge-intensive and commonsense benchmarks, with
ARC (JA) improving by $+8.52$ points, JCommonsenseQA~\citep{kurihara2022jglue}
by $+0.8$ points, and Japanese History ~\citep{jmmlu2023} by $+1.8$ points. However, mathematical
reasoning degrades sharply, with GSM8K (JA)~\citep{cobbe2021training} falling
$9.3$ points relative to the Qwen3-Instruct baseline. The overall average remains
nearly unchanged (75.93 $\to$ 76.38), masking a pronounced within-task
redistribution: commonsense and factual recall improve while multi-step reasoning
erodes. This pattern is precisely the diagnostic signature observed in the Hebrew
instruct result in Table~\ref{tab:hebrew_instruct}, where SFT without prior CPT
left deep-layer routing collapse unremediated and the model redistributed
competencies rather than broadening them. As in the Hebrew case, the Shisa entropy
and Gini profiles converge to the base model in the final layers
(Figures~\ref{fig:japanese}b and~\ref{fig:japanese}d), strengthening the case that the
performance redistribution is a direct consequence of unremediated deep-layer
routing collapse rather than a training artifact specific to either language.
 
\begin{table}[h]
\centering
\caption{%
  Japanese benchmark results for SFT model variants.
  Higher is better for all metrics.
  Avg = macro-average across the four tasks.
  GSM8K \citep{cobbe2021training};
  ARC \citep{clark2018think};
  JCommonsenseQA \citep{kurihara2022jglue}.
  All benchmarks evaluated by the authors.%
}
\label{tab:japanese_instruct}
\resizebox{\textwidth}{!}{%
\begin{tabular}{llccccc}
\toprule
\textbf{Model} & \textbf{Training} &
  \textbf{GSM8K (JA)} &
  \textbf{ARC (JA)} &
  \textbf{Japanese History (JA)} &
  \textbf{JCommonsenseQA} &
  \textbf{Avg} \\
\midrule
Qwen3-30B-A3B (instruct) & SFT (original) & \textbf{86.8} & 62.00          & 62.2          & 92.7           & 75.93 \\
Shisa-Qwen3-30B          & SFT            & 77.5          & \textbf{70.52} & \textbf{64.00} & \textbf{93.5}  & \textbf{76.38} \\
\bottomrule
\end{tabular}%
}
\end{table}
\paragraph{Correspondence to the Hebrew leaderboard.}
 
Taken together, Tables~\ref{tab:japanese_base} and~\ref{tab:japanese_instruct} replicate,
in Japanese, the two-tier pattern first documented for Hebrew in
Tables~\ref{tab:hebrew_base} and~\ref{tab:hebrew_instruct}.
In both languages, CPT produces the most consistent and broadest performance gains, particularly on tasks requiring rich linguistic and cross-lingual knowledge, whereas SFT alone yields a highly uneven profile characterized by large swings in opposite directions: GSM8K \citep{cobbe2021training} rises from 60.0 to 72.0 in Hebrew while HellaSwag \citep{zellers2019hellaswag} drops from 70.0 to 52.0, and GSM8K \citep{cobbe2021training} falls 9.3 points in Japanese, reflecting a redistribution of competencies rather than clear evidence of a broadening of linguistic capacity.
This correspondence is consistent with the routing dynamics reported in
Section~\ref{sec:japanese}: just as Hebrew CPT substantially raised usage entropy,
reduced Gini inequality, and increased the share of language-agnostic experts (Section~\ref{sec:specialization}),
Japanese CPT (Swallow) achieves the same structural correction and the same downstream
pattern.
Conversely, just as Hebrew SFT (instruct64) produced a less substantial improvement
in expert distribution balance compared to the Nemotron base256 CPT, Shisa's routing profiles similarly
show more limited gains (compared to Swallow), and its benchmark profile reflects a redistribution rather
than a broadening of competencies.
The parallel across two typologically unrelated languages — differing in script, morphology,
and syntactic structure — supports the interpretation that the performance consequences
of routing imbalance may represent a recurring pattern of under-resourced
pre-training in MoE systems, and not an artifact of any particular language's
linguistic properties.
\section{Discussion and Conclusion}
\label{sec:discussion}

\subsection{Summary of Findings}

While concurrent work has begun to characterize cross-lingual routing dynamics in MoE models \citep{bandarkar2026}, this work presents, to our knowledge, the first systematic comparison of the routing advantages conferred by CPT versus SFT for underrepresented languages, and the downstream performance consequences of each. By analyzing two architecturally distinct MoE models — a pure Transformer (Qwen3-30B-A3B) and a hybrid Mamba-Transformer (Nemotron-3-Nano-30B-A3B) — using a suite of complementary routing metrics, we have identified several consistent phenomena that illuminate how MoE architectures internally allocate computational capacity across languages.

Our central observation is a clear correspondence between improved downstream task performance and improved routing dynamics: models that achieved significant gains on Hebrew and Japanese benchmarks after training consistently exhibited higher usage entropy, more uniform expert distribution, and a shift from language-specific experts toward shared bilingual routing. This link between benchmark improvement and internal routing reorganization suggests that routing metrics offer a principled diagnostic lens for multilingual capacity in MoE systems.

Five key findings support and extend this observation:

\begin{enumerate}

\item \textbf{Deep-layer routing collapse in underrepresented languages.} In both pre-trained base models, the final layers exhibit a sharp narrowing of expert utilization for Hebrew tokens, while English maintains broad, uniform routing. This collapse manifests consistently across all metrics as reduced usage entropy, elevated Gini coefficients, and decreased cosine similarity, providing evidence that it is a robust pattern across the routing metrics analyzed rather than an artifact of any single metric \citep{fedus2021, zoph2022}. Fine-tuning on data with greater Hebrew representation substantially mitigates this collapse across all model variants, with the Nemotron model family showing more complete correction than Qwen in the experiments reported here.

\item \textbf{Architectural differences in routing patterns.} The two base models exhibit distinct routing behaviors. The hybrid Nemotron model shows greater layer-to-layer variability, likely attributable to the interleaving of MoE layers with Mamba-2 and attention layers but maintains higher overall entropy values. The pure Transformer Qwen model produces more stable routing across layers but with lower entropy and a higher proportion of language-specific experts \citep{fedus2021, lepikhin2020}.

\item \textbf{Cross-linguistic generalization of routing collapse.} Both Hebrew and Japanese, two typologically unrelated languages, differing in script, morphology, syntactic structure, and writing direction exhibit similar deep-layer collapse signatures in their respective base models. Training on data that better represents these languages produces similar improvements in both cases: increased entropy, reduced Gini inequality, and a shift toward bilingual expert sharing. Furthermore, CPT appears substantially more effective at correcting routing imbalance than SFT alone.

\item \textbf{Hybrid architectures may offer greater routing correctability.} The Nemotron model demonstrates both larger benchmark improvements and more thorough routing correction after fine-tuning in the experiments reported here: experts become nearly entirely bilingual, and distribution increases substantially across all layers. This more complete reorganization, potentially facilitated by the complementary representational capacity of the Mamba-2 state space layers \citep{dao2024, gu2023}, suggests that hybrid architectures may be more amenable to adapting MoE routing for underrepresented languages than pure Transformer designs. However, this conclusion remains suggestive because the models differ in pre-training data, routing configuration, shared-expert design, and number of MoE layers.

\end{enumerate}
\subsection{Interpretation: Why Do Deep Layers Collapse?}
\label{sec:interpretation}

The consistent localization of routing collapse in the deepest layers, across both architectures and all metrics, invites a mechanistic interpretation. In autoregressive language models, the final layers are responsible for the most abstract, language-specific computations, transforming shared semantic representations into language-specific lexical predictions \citep{tenney2019, conneau2020}. These layers must map from a shared semantic latent space to a language-specific distribution over the output vocabulary.

For a high-resource language like English, which dominates the pre-training corpus, the model has had ample opportunity to develop diverse routing pathways through these final layers, distributing the computational load across many experts. For Hebrew, which constitutes a small fraction of the pre-training data, the model has encountered far fewer diverse contexts requiring Hebrew-specific output computation \citep{lauscher2020}. As a result, the router learns to funnel Hebrew tokens through a narrow, repeatedly activated subset of experts in the deep layers. This is a form of \textit{routing path dependence} in which limited pre-training exposure constrains the diversity of learned routing solutions \citep{fedus2021, lepikhin2020}.

This interpretation is further supported by the behavior of middle layers, which perform more language-agnostic semantic processing and exhibit the highest cross-lingual similarity across all models (as shown in Supplementary Notes, measured by cosine similarity, Top-K overlap, and rank correlation ) \citep{conneau2020}. The middle layers appear to operate as a shared computational substrate for both languages, while the boundary layers (early and deep) handle language-specific processing: early layers process surface-level features (script, tokenization, morphological cues), and deep layers generate language-specific output representations \citep{tenney2019}.
The replication of this pattern in Japanese, a language typologically unrelated to Hebrew, is consistent with this interpretation, suggesting that the collapse mechanism is driven by token frequency statistics in pre-training rather than by any language-intrinsic property.

\subsection{Implications for Low-Resource Language Adaptation}
\label{sec:implications}

Our analysis suggests four strategic priorities for pre-training underrepresented language adaptation of MoE architectures:

\paragraph{Routing metrics as diagnostics.} The suite of metrics introduced in this work: usage entropy, Gini coefficient, LSI, cosine similarity, selection entropy, Top-K overlap, and rank correlation (See Supplementary Notes), provides an interpretable and computationally inexpensive diagnostic toolkit for assessing multilingual capacity allocation in MoE models \citep{fedus2021}. Rather than relying solely on downstream task performance (which conflates many factors), practitioners can use routing analysis to identify specific layers and experts where a target language is underserved, enabling more precise corrective action.

\paragraph{Balanced bilingual fine-tuning is effective.} Our results demonstrate that fine-tuning on a corpus with a deliberate 60\% Hebrew / 40\% English language ratio substantially corrects routing imbalance, even with a relatively modest token budget compared to the pre-training. This suggests that routing reorganization is a relatively plastic process that does not require retraining from scratch, making it accessible even under constrained computational budgets \citep{gururangan2020}.

\paragraph{Architecture may matter for correctability.} For practitioners, the present results suggest while not proving that hybrid Mamba-Transformer MoE designs may provide a more forgiving substrate for underrepresented-language adaptation. The superior correctability of the hybrid Mamba-Transformer architecture suggests that state-space components may offer a more plastic substrate for reorganizing routing patterns during adaptation, potentially through the complementary representational capacity of state space layers providing additional degrees of freedom during fine-tuning \citep{dao2024, gu2023}. This advantage holds consistently across both training strategies tested on Hebrew, where Nemotron largely eliminates deep-layer collapse while CPT Qwen3 retains residual imbalance, and is broadly supported by the Japanese results on Qwen3. Consistently with this, Nemotron base256 achieves stronger downstream benchmark gains on Hebrew than the CPT Qwen3 model, suggesting that more complete routing reorganization is associated with broader linguistic competence.

\paragraph{CPT appears more reliable than SFT for routing correction.} Both CPT and direct SFT partially correct routing imbalance, but CPT produces a substantially more thorough reorganization. For Nemotron, CPT (base256) largely eliminates deep-layer collapse and drives expert utilization toward near-complete language agnosticism. SFT (instruct64) produces meaningful but less complete routing corrections, with correspondingly uneven benchmark gains. For Qwen3, CPT raises routing entropy and reduces Gini inequality across most layers, though a residual deep-layer imbalance persists, possibly reflecting lower routing plasticity in the pure Transformer setting studied here. SFT alone failed to remediate Qwen's deepest layers. However, in Nemotron, it did resolve the collapse, though CPT still proved much more effective. For practitioners developing LLMs for underrepresented languages, the additional computational investment in continual pre-training may be justified, though SFT remains a viable option under tighter computational constraints \citep{gururangan2020}.

\subsection{Limitations}
\label{sec:limitations}

Despite these insights, several limitations remain. While our analysis spans two typologically distant languages (Hebrew and Japanese), the routing collapse phenomenon and the CPT-versus-SFT comparison are established on different model families: the Hebrew analysis spans two architectures with partially confounded training strategies, and the controlled CPT-versus-SFT comparison is restricted to a single architecture (Qwen3-30B-A3B) and a single non-English language. Robustness to other languages, architectures, and resource levels remains to be established. Second, we analyze only two model families, both at the $\sim$30B parameter scale with $\sim$3B active parameters; routing dynamics may differ at other scales or with different expert counts and top-K configurations. Third, our evaluation dataset (TED talk transcriptions) represents a specific domain of natural discourse, and routing behavior may vary across other domains (e.g., technical, literary, or conversational text). Future work could explore data-driven approaches to threshold selection or continuous measures of specialization. Fourth, tokenization differences across Hebrew, Japanese, and English may affect token-level routing statistics; future work should explicitly control for segmentation effects. Fifth, all routing measurements are based on TED-style discourse, and broader domain coverage is needed before claiming domain-general behavior.

\subsection{Concluding Remarks}
\label{sec:concluding_remarks}

This study demonstrates that internal routing dynamics in MoE language models encode measurable and interpretable multilingual biases that mirror the data imbalances of pre-training \citep{bender2021, conneau2020}. The deep-layer routing collapse we identify for Hebrew, a morphologically rich low-resource language, represents a previously undocumented bottleneck that constrains the model's capacity to process underrepresented languages. Critically, this bottleneck is not immutable. Balanced bilingual fine-tuning substantially reorganizes routing patterns, broadening expert utilization and driving the system toward language-agnostic computation \citep{fedus2021, gururangan2020}. These routing improvements are associated with downstream performance improvements, with CPT-trained models achieving consistent Hebrew and Japanese benchmark gains that SFT-only models fail to match. The degree of correction depends on both the architectural substrate and the training strategy, with the hybrid Mamba-Transformer model appearing more amenable in our experiments, subject to the architectural and pre-training confounds discussed above.

More broadly, our results establish expert routing analysis as a principled and practical lens for understanding multilingual capacity allocation in MoE systems. By moving beyond aggregate performance metrics to examine the internal computational pathways through which different languages are processed, we gain a richer understanding of how these models succeed and fail in serving the world's linguistic diversity. The replication of the collapse signature in Japanese, a language typologically unrelated to Hebrew in script, morphology, and syntax, provides initial cross-linguistic evidence that this bottleneck is not specific to any single language family but rather a recurring consequence consistent with pre-training data imbalance. As MoE architectures continue to grow in scale and deployment, ensuring equitable routing across languages will be essential for building truly multilingual AI systems.

\bibliographystyle{plainnat}
\bibliography{references}

\end{document}